\pdfoutput=1

\documentclass[11pt]{article}

\usepackage[]{acl}
\usepackage{times}
\usepackage{latexsym}
\usepackage[T1]{fontenc}
\usepackage[utf8]{inputenc}
\usepackage{microtype}
\usepackage{inconsolata}
\usepackage{graphicx}
\usepackage{multirow}
\usepackage{enumitem}
\setenumerate[1]{label=(\arabic*)}

\newcommand\satyrn{\textsc{Satyrn}}

\title{\satyrn{}: A Platform for Analytics Augmented Generation}

\author{
    \quad Marko Sterbentz
    \quad Cameron Barrie
    \quad Shubham Shahi
    \quad Abhratanu Dutta\\
    \quad {\bf Donna Hooshmand}
    \quad {\bf Harper Pack}
    \quad {\bf Kristian J. Hammond}\\
    Northwestern University\\
    {\tt \{marko.sterbentz, cameron.barrie\}@u.northwestern.edu} \\
    {\tt kristian.hammond@northwestern.edu}
}

\begin{document}
\maketitle

\begin{abstract}
Large language models (LLMs) are capable of producing documents, and retrieval augmented generation (RAG) has shown itself to be a powerful method for improving accuracy without sacrificing fluency. However, not all information can be retrieved from text. We propose an approach that uses the analysis of structured data to generate fact sets that are used to guide generation in much the same way that retrieved documents are used in RAG. This analytics augmented generation (AAG) approach supports the ability to utilize standard analytic techniques to generate facts that are then converted to text and passed to an LLM. We present a neurosymbolic platform, \satyrn{}, that leverages AAG to produce accurate, fluent, and coherent reports grounded in large scale databases. In our experiments, we find that \satyrn{} generates reports in which over 86\% of claims are accurate while maintaining high levels of fluency and coherence, even when using smaller language models such as Mistral-7B, as compared to GPT-4 Code Interpreter in which just 57\% of claims are accurate.
\end{abstract}

\section{Introduction}

\begin{figure*}[ht]
\centering 
\includegraphics[width=\linewidth,keepaspectratio]{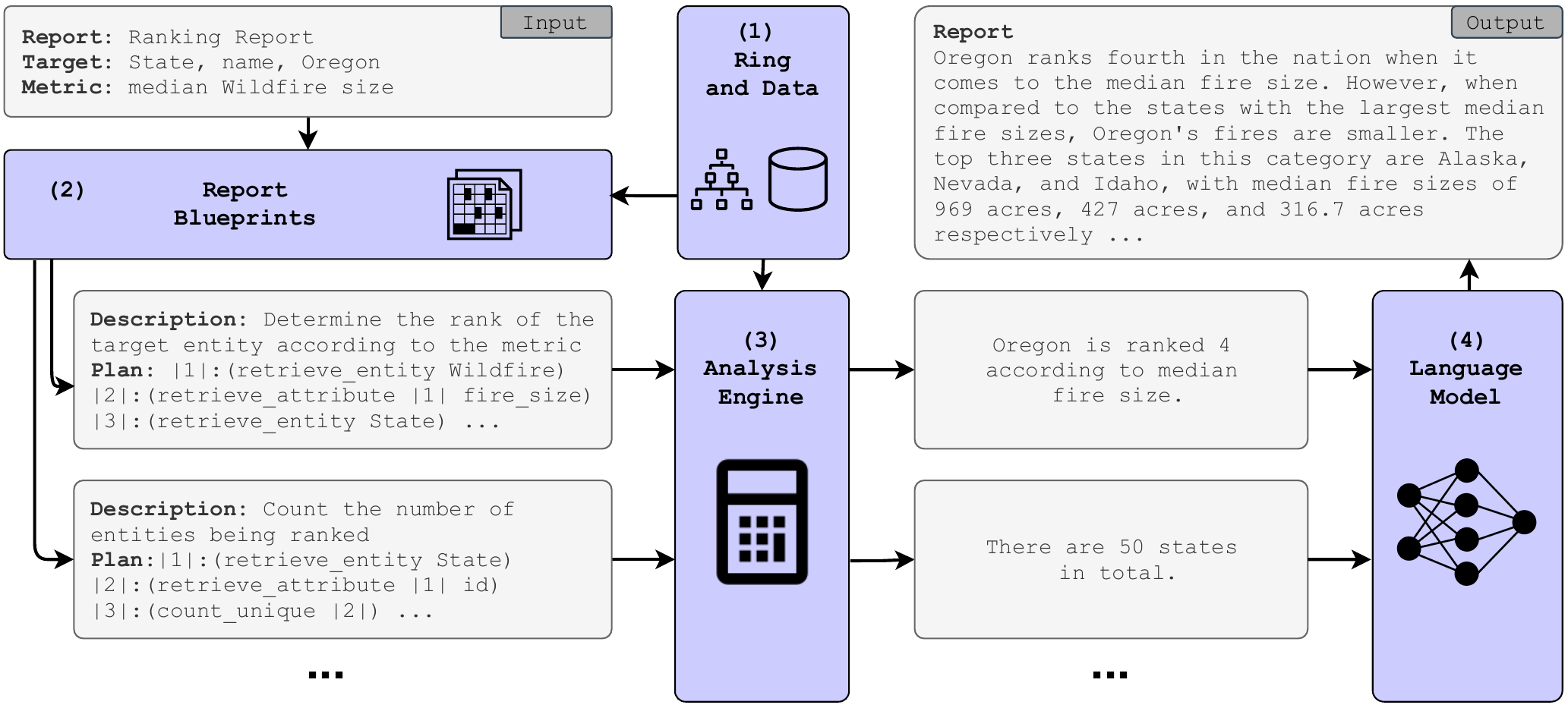}
\caption{The high level approach of \satyrn{} and its analytics augmented generation.} 
\label{fig:satyrn_diagram} 
\end{figure*}

In recent times, large language models (LLMs) have seen a meteoric rise in popularity due to their ability to generate fluent and coherent language. However, these models often struggle to produce truthful outputs \cite{Ji2022SurveyOH, huang2023survey, borji2023categorical}. Methods such as retrieval augmented generation (RAG) \cite{lewis2021retrievalaugmented} have been developed to address such concerns. Yet, they too have an important limitation: they only work when the required information is present in textual form. For many sources of information, such as relational databases, the sought after information must be derived via computation and analysis of those data. Performing this analysis and providing the results to an LLM is another way to augment their generation with a far broader array of information. Such analytics augmented generation (AAG) supports the use of standard analytics techniques to generate facts that are converted to text and used to guide the generation of accurate responses.

This has motivated the development of methods in which an LLM invokes tools and external computation engines for deriving information from structured data \cite{schick2024toolformer, masterman2024landscape}. This is also the approach of OpenAI's Code Interpreter\footnote{https://platform.openai.com/docs/assistants/tools/code-interpreter} which can be used for generating reports that present information derived from structured data. However, these approaches let the LLMs decide when to call these tools and write the queries required to derive the information, an inherently non-deterministic procedure that can make them unreliable in producing the set of information to communicate in a report.

In this paper, we present \satyrn{}\footnote{https://github.com/nu-c3lab/satyrn}, a neurosymbolic platform which leverages AAG for producing factual, fluent, and coherent reports. \satyrn{} separates the analysis from the language generation process in order to steer LLMs using facts derived from relational databases. It deterministically produces the analysis plans required to satisfy the informational needs of a report, generates templated language describing the results, and \textit{then} uses a language model to generate the final report. This separation of the analysis and generation steps along with the deterministic nature of the analysis planning improves factual accuracy compared to existing solutions like Code Interpreter. In our experiments on the report generation task, we see a factual accuracy of 86\% using \satyrn{} compared to 57\% using Code Interpreter.

One of \satyrn{}'s highlights is its ability to effectively utilize small, off-the-shelf models to achieve highly accurate results. In our experiments, the two local models we used were Mistral-7B and 4-bit quantized Mixtral-8x7B, each running on a single A6000 GPU. This showcases the ability of running \satyrn{} locally under resource constraints.

Our main contributions are as follows:
\begin{itemize}
    \item First, we propose \satyrn{}, a scalable, domain-agnostic, and neurosymbolic platform for producing fluent and coherent reports that adhere to the source data (\S \ref{sec:method}).
    \item Second, we demonstrate how \satyrn{} can be used for producing reports and compare it with other methods (\S \ref{sec:experiments}). In our experiments, we generate 200 reports with over 3200 claims, across 3 broadly applicable report types and 8 domains and evaluate the reports for accuracy, fluency, and coherence. 
    \item Third, we provide an analysis of the factual accuracy, fluency, and coherence of generated reports (\S \ref{sec:results}). We find that reports produced by \satyrn{} strongly adhere to the data while maintaining high degrees of fluency and coherence, even when using smaller models.
\end{itemize}
\section{Methods}
\label{sec:method}

In this section, we present the components of \satyrn{} that allow it to perform analytics on the data, irrespective of its domain, in order to generate accurate, fluent, and coherent reports grounded by this data. The high level architecture of \satyrn{} is shown in Figure \ref{fig:satyrn_diagram}. Beginning with the data in (1), we create a lightweight knowledge representation, called a \textit{ring}, that provides simple semantic labels that describe how the data can be analyzed by \satyrn{}. We introduce a new plan representation language called Structured Question Representation (SQR) (pronounced \textit{seeker}) that allows complex analyses to be specified according to the semantic enhancements.

The input to \satyrn{} is a structured information request and the output is a report satisfying this request with an answer and contextualizing information. This request is a JSON object describing the type of report to generate, the subject of the report, a metric to use when evaluating the subject, and any filters to apply to the data before analysis. Report blueprints (2) specify the information to be derived via analysis of the data and are used to produce a set of executable plans. With the analysis engine (3), each of these plans is executed in order to derive the information. The outputs are formatted as statements and used as input to the LLM (4) which generates the final report.

\subsection{\satyrn{} Rings}

\satyrn{} requires knowledge of the objects the data describes in order to apply analytics and produce information regardless of a dataset's domain.

\begin{figure}[!t]
\centering 
\includegraphics[width=\columnwidth]{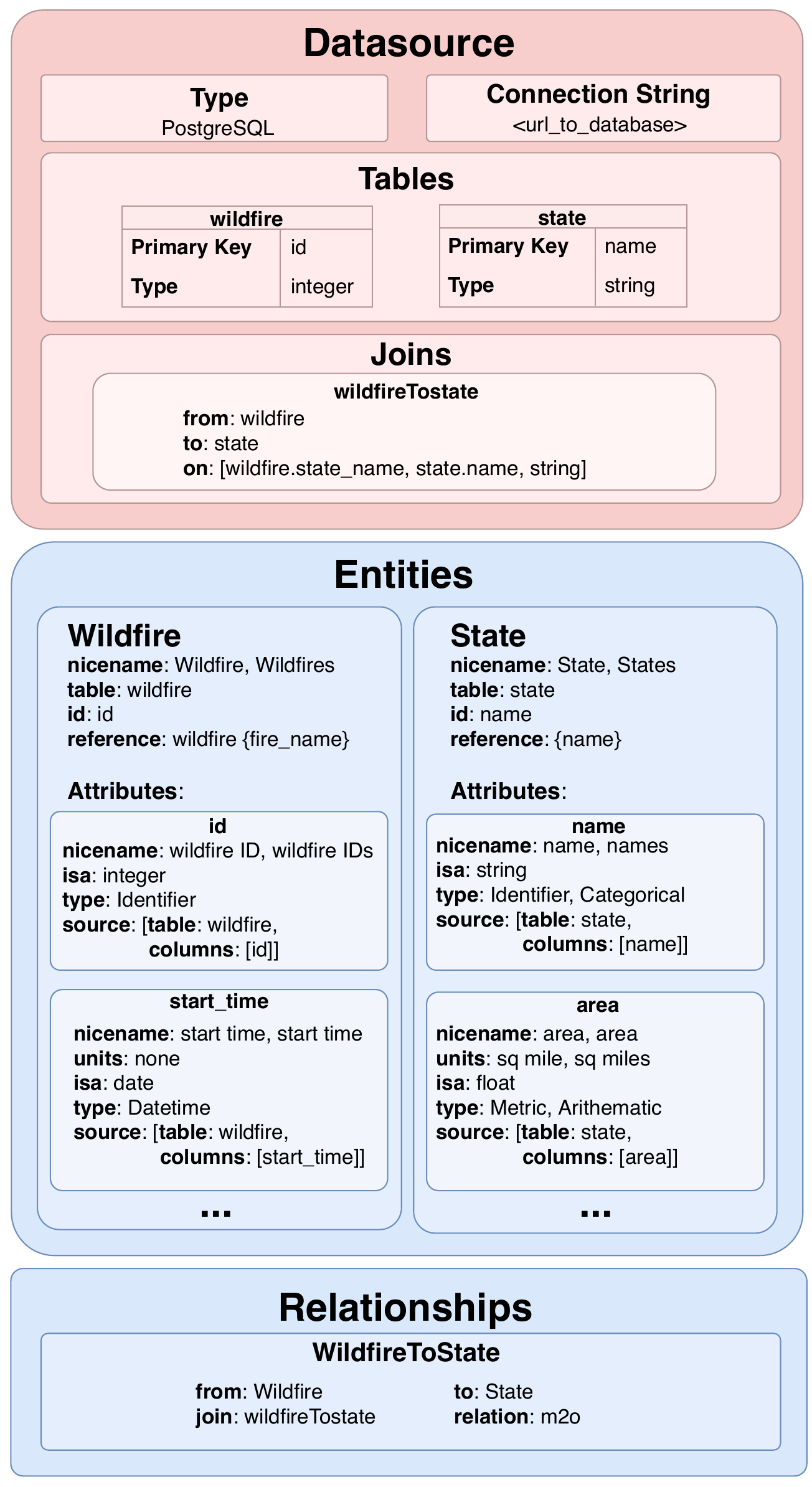}
\caption{An example of a \satyrn{} ring with two entities defined: State and Wildfire.} 
\label{fig:satyrn_ring} 
\end{figure}

To achieve this, we define a lightweight labeling called a \textit{ring} which specifies the entities in the data, their attributes, and the relationships between these entities. This labeling allows \satyrn{} to run analyses on the data without the need for any domain information. New datasets can be utilized within \satyrn{} by creating a ring and filling in the necessary details described below. Figure \ref{fig:satyrn_ring} depicts an example of the makeup of a ring.

The main definitions in a \satyrn{} ring are:
\begin{itemize}
\item \textbf{Entities}: objects of interest described by the data, such as a person, place, or thing. For example, in Figure \ref{fig:satyrn_ring}, there are entities for Wildfire and State.
\item \textbf{Attributes}: properties of an entity. Each entity has one or more attributes that map to columns in the underlying database tables.

\item \textbf{Relationships}: the connection between two entities in the ring. 

\end{itemize}

In order for \satyrn{} to decide how to apply analytics to attributes, it needs a finer grained characterization of them than is provided by the database. For instance, performing arithmetic operations on an attribute requires it to be numerical. To capture this, we define six primary attribute types: 

\begin{itemize}
    \item \textit{Arithmetic}: numerical values for which mathematical operations make sense.
    \item \textit{Categorical}: discrete values typically denoting a class or category.
    \item \textit{Datetime}: values designating a date and time. 
    \item \textit{Document}: values containing free text.
    \item \textit{Identifier}: values meant to be used as a unique identifier for an entity.
    \item \textit{Metric}: values meant to be used as a measure for ranking and comparing entities.
\end{itemize}

Rings also contain domain-specific information. We define \textit{nicenames} for both attributes and entities to provide more descriptive labels than column headers. Units (e.g., dollars, acres) can also be defined for attributes. This domain knowledge is used to help \satyrn{} better express the information it derives in natural language.

The knowledge encapsulated by a ring makes it simple to specify and execute analytic operations. For instance, attributes of an entity can be present in multiple underlying data tables necessitating joins between these tables when retrieving these values. We define these joins inside the ring, abstracting away the need to specify them at runtime when performing operations that span multiple database tables. Similarly, when \satyrn{} performs an analytic operation involving multiple entities, joins are needed to connect the underlying database tables. For each pair of entities, relationships encapsulate the required joins so these details can be abstracted away during the specification of the analytics.

\subsection{Structured Question Representation (SQR)}

In order to leverage the knowledge in a ring and effectively apply analytics, an expressive and compositional plan representation in which plans can be reused across datasets is required. To satisfy these representational needs, we define an analytic plan representation language, SQR, which abstracts away specific implementation details of the underlying query language. This simplifies the syntax and makes it agnostic to the data storage format and corresponding query language. SQR allows for the specification of plans in which the entities and attributes defined in the ring are retrieved and analyzed using analytic operations.

\begin{figure}[!h]
\centering 
\includegraphics[width=\columnwidth]{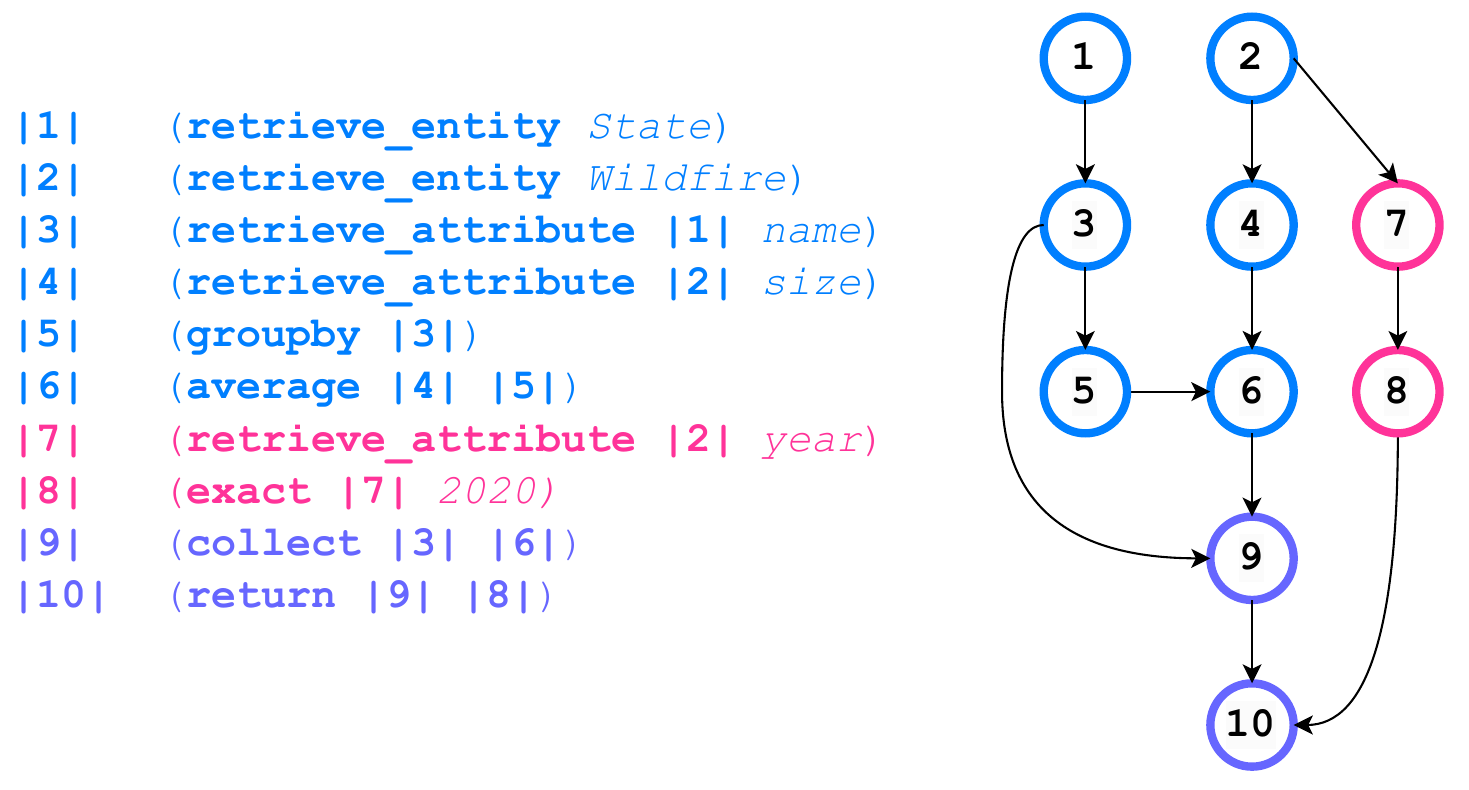}
\caption{A SQR plan, in textual and graph forms, for determining average wildfire size for each state in 2020.}
\label{fig:sqr_example} 
\end{figure}

A SQR plan is represented as a directed acyclic graph that specifies an ordered series of steps to carry out, wherein operations are chained together in order to retrieve and analyze data. Each node of the graph represents an analytic operation whose output is fed to later steps that require the result. The inputs and outputs of these analytic operations utilize attribute types in order to determine how operations can be chained together to form a complete SQR plan. Existing operations enable aggregation (e.g., average, count, sum), comparison (e.g., greater than, exact, not), and data manipulation (e.g., sort, groupby, limit). A full listing of the SQR operations that are currently implemented within \satyrn{} is provided in Appendix \ref{app:sqr_operations}. New operations can be added to incorporate new analytic capabilities. Arbitrarily complex plans can be composed to satisfy any information goal that can be described with the available data and analytics. A SQR plan example can be seen in Figure \ref{fig:sqr_example}.

Query languages like SQL require joins between tables to be explicit in the query. However, joins are not needed within SQR. Any required joins are encoded within the ring and are used in the definition of relationships between entities. 
Such joins are automatically added to the final query against the database when parsing SQR to SQL. 

\subsection{SQR Plan Templates}
\label{sec:plan_templates}

A SQR plan template is a SQR plan with slots to be filled with entities, attributes, values, and operations. All slots used within the SQR plan template correspond to the attribute types used in a ring, making the plan templates domain-agnostic and thus reusable with any ring. We define a set of SQR plan templates for use within \satyrn{}. Each SQR plan template has an associated statement template that describes the result of executing the plan template. These statement templates have slots for the natural language expression of specific entities, attributes and values, that correspond to the slots in the SQR plan templates.

\subsection{Analysis Engine}

SQR plans first need to be converted to a query format that is native to the data source (e.g., SQL for relational databases) in order to be executed to derive results. We build an analysis engine that is capable of parsing SQR plans, converting them to valid SQL queries against the underlying data, and producing the \textit{nicenames} and units for all query results. Implementation details are provided in Appendix \ref{app:analytics_engine}.

\subsubsection{Generation of Factual Statements}
\label{sec:statement_generation}
The output of the SQL query derived from a SQR plan is a set of raw result tuples that lack contextual information, making it difficult for an LLM to parse. To address this issue, the analysis engine can produce statements in natural language by utilizing the statement template associated with a SQR plan template and filling its slots with the values, \textit{nicenames}, and units obtained by executing a SQR plan. Examples of generated statements can be seen in Figure \ref{fig:satyrn_diagram} as the output of the Analysis Engine.

\subsection{Domain-Agnostic Report Blueprints}

\begin{figure*}[ht]
\centering 
\includegraphics[width=\linewidth,keepaspectratio]{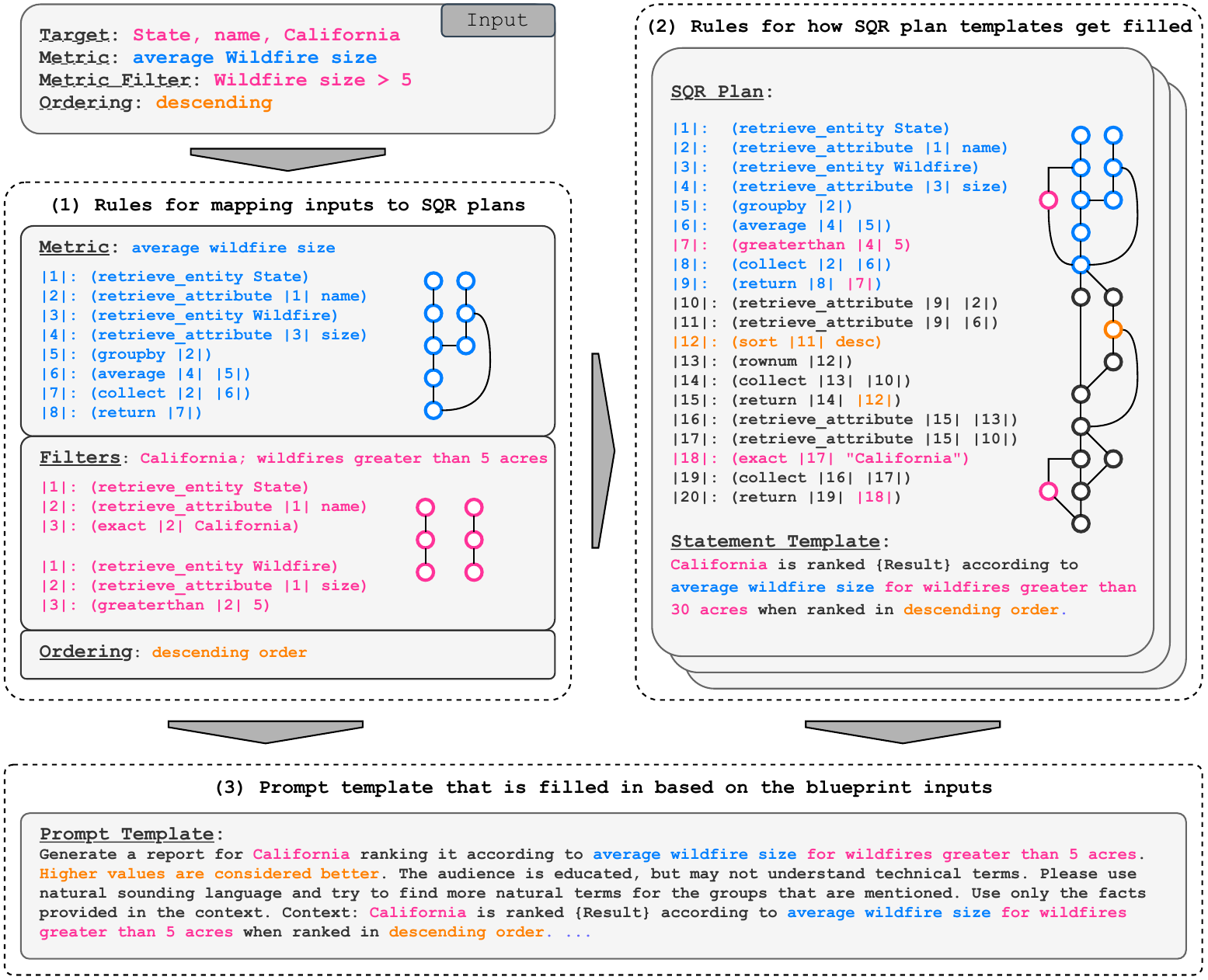}
\caption{An instantiation of a report blueprint for ranking California against other states by average wildfire size. The resulting SQR plans will be executed and their results are inserted into the prompt to form the input to the LLM.} 
\label{fig:blueprint_diagram} 
\end{figure*}

For each kind of report, we create a report blueprint that defines the information that should be included in the report. Each blueprint comprises a collection of SQR plan templates that define the core analytics to perform in order to satisfy the information goals of this type of report. Inputs to a blueprint include the target entity, the metric of interest, and any filters that should be used to retrieve and filter the data for analysis. Additional inputs can also be defined as required.

Blueprints provide three key elements required for generating reports:
\begin{enumerate}
    \item \textbf{Rules for mapping inputs to SQR plans}. Rules map the target entity and any associated filters provided by the user to SQR plans for retrieving and filtering the data. For the metric, there is a rule that produces the SQR plan for deriving the value of this metric from the data.
    
    \item \textbf{Rules for how SQR plan templates get filled}. The SQR plans and filters that were produced with the rules in (1) are assigned to each SQR plan template according to the requirements of that plan template and the information goals of the report. Using these assignments, each SQR plan template is filled with the proper entities, attributes, and filters to produce an executable SQR plan.

    \item \textbf{A prompt template that is filled in based on the blueprint inputs}. This includes a description of the report to be generated, and the set of factual statements generated by \satyrn{}.

\end{enumerate}

An example blueprint which shows how these three key elements are used for a specific user input can be seen in Figure \ref{fig:blueprint_diagram}. The blueprint rules make use of the attribute types defined in the ring. This allows inputs from the users to be mapped to plans irrespective of any domain specifics. As a result, blueprints scope across any domain and are reusable for any ring which is provided to \satyrn{}. Implementation details for the blueprints are provided in Appendix \ref{app:blueprint_implementation}.

\subsection{Execution of a Report Blueprint} Blueprint execution requires an input of a target entity, relevant metric information, and filters that are to be applied. Using the defined rules, the plan templates in the blueprint gets populated to provide executable SQR plans. The analysis engine executes these plans and generates language describing the results using the statement templates associated with the plan templates. Finally, the prompt template defined in the blueprint gets populated with the generated factual statements and is sent to the language model to generate a coherent, fluent, and factual report. Example report outputs can be seen in Appendix \ref{app:example_docs}.
\section{Experiments}
\label{sec:experiments}

We evaluate the performance of \satyrn{} along with two baselines on the task of generating reports for a specific information request. Reports are evaluated based on their accuracy, fluency, and coherence. We create three report blueprints: a \textbf{Ranking} report that ranks an entity's performance amongst its cohort, a \textbf{Time over Time} report that compares an entity's performance at two time periods, and a \textbf{Comparative Benchmark} report that compares an entity's performance with a benchmark value. Further details on the specific information requirements of each report type can be found in Appendix \ref{app:report_types}. These three report blueprints are each used across eight domains: healthcare, environmental sustainability, urban housing, criminal justice, education, legal and judicial, socioeconomic, and business. A summary of the datasets can be found in Appendix \ref{app:datasets}. We create a ring for each dataset.

\subsection{Report Generation Modes}

\textbf{\satyrn{}}: This generation mode utilizes \satyrn{} for generating the reports. It uses the factual statement generation detailed in Section \ref{sec:statement_generation} when building the prompt. We use three LLMs for generation: Mistral-7B \cite{jiang2023mistral}, Mixtral-8x7B \cite{jiang2024mixtral}, and GPT-4. Mixtral-8x7B was quantized to 4-bits \cite{frantar2023gptqaccurateposttrainingquantization}.

\textbf{\satyrn{}-Table (ablation)}: This mode is the same as \satyrn{}, except that factual statements are replaced by tables containing the results of executing the underlying SQL query. This is an ablation for determining the utility of the factual statements versus raw SQL results as prompting inputs.

\textbf{Unaugmented GPT-4 (baseline)}: The first baseline generation mode tests the model's ability to use its parametric knowledge to generate a factual report. Instead of augmenting the generation with information derived via analysis, we provide the target entity, metric and a description of the information to be included in the report. 

\textbf{Code Interpreter (baseline)}: We also compare \satyrn{} with OpenAI’s Code Interpreter which can take one or more data files, write code to extract information, and then generate natural language outputs. We provide the model with descriptions of the information to present in the prompt. Given the data and the prompt, it must determine how to derive this information and generate a report. We give the tool direct access to the same data used by \satyrn{}. However, due to limitations on the number of files and their sizes, we cannot generate reports for domains with larger datasets, namely the legal and judicial, environmental sustainability, and business domains. Thus, we generate reports with Code Interpreter for only five domains.

\begin{table*}[ht!]
\small
\centering
\begin{tabular}{p{0.16\linewidth}p{0.12\linewidth}p{0.10\linewidth}p{0.10\linewidth}p{0.13\linewidth}p{0.09\linewidth}p{0.11\linewidth}}
\hline

\multirow{2}{\linewidth}{\textbf{Generation Mode}}
& \multirow{2}{\linewidth}{\textbf{Model}}
& \textbf{Fraction Factual} $\uparrow$ 
& \textbf{Fraction Refuted} $\downarrow$ 
& \multirow{2}{\linewidth}{\textbf{Fraction Confabulated} $\downarrow$} 
& \multirow{2}{\linewidth}{\textbf{Fluency} $\uparrow$}
& \multirow{2}{\linewidth}{\textbf{Coherence} $\uparrow$} \\
\hline

Unaugmented & GPT-4 & 0.477 & 0.108 & 0.416 & 0.928 & 0.971 \\
\hline

Code Interpreter* & GPT-4 & 0.570 & 0.337 & 0.093 & 0.904 & 0.919 \\
\hline

\multirow{3}{\linewidth}{\satyrn{}-Table}
& Mistral-7B & 0.548 & 0.314 & 0.138 & 0.887 & 0.946 \\
& Mixtral-8x7B & 0.628 & 0.194 & 0.178 & 0.890 & 0.970 \\
& GPT-4 & 0.832 & 0.032 & 0.136 & 0.944 & 0.979 \\
\hline

\multirow{3}{\linewidth}{\satyrn{}}
& Mistral-7B & \textbf{0.891} & 0.038 & \textbf{0.071} & 0.887 & \textbf{0.980} \\
& Mixtral-8x7B & 0.863 & 0.034 & 0.103 & 0.878 & 0.929 \\
& GPT-4 & 0.863 & \textbf{0.011} & 0.126 & \textbf{0.946} & 0.977 \\

\hline

\end{tabular}
\caption{Average accuracy, fluency, and coherence scores for all models and generation modes used in the evaluation. 
*Code Interpreter only evaluated on five of the eight domains due to data file size constraints.}
\label{tab:accuracy_fluency_coherence}
\end{table*}

\subsection{Report Evaluation}

We use three metrics for evaluating reports: fluency, coherence, and accuracy. We evaluate the first two using the \texttt{summarization} checkpoint of UniEval \cite{zhong-etal-2022-towards}. The source documents used in the coherence scoring are the factual statements associated with that report. For determining the accuracy, we manually examine each report to find the percentage of claims made that are supported by facts derived from the data. The ground truth facts are produced by \satyrn{} and used when evaluating across all modes. Discrepancies in the reports generated by the baselines and the facts generated by \satyrn{} were manually checked by examining the data to ensure fairness. The accuracy evaluation was carried out in two steps: claim identification and claim classification. Each claim is classified as either \textbf{factual} (supported by the data), \textbf{refuted} (contradicted by the data), or \textbf{confabulated} (not inferrable from the data). For full class definitions and the rubric used for identifying and classifying claims, see Appendix \ref{app:accuracy_evaluation}. Each report was evaluated by two of the authors, and we computed an inter-annotator agreement score with Krippendorff's $\alpha$ \cite{hayes2007answering} using the ratio of factual to total claims, resulting in an $\alpha$ of 0.86.
\section{Results}
\label{sec:results}

In terms of factual accuracy, \satyrn{} outperforms the closest baseline by 29 points, even with smaller models, while maintaining similarly high degrees of fluency and coherence as measured by UniEval. These primary results are captured by Table \ref{tab:accuracy_fluency_coherence}.

\subsection{Factual Accuracy}
\begin{figure} % Add * after figure to make it span two columns
\centering 
\includegraphics[width=\columnwidth,keepaspectratio]{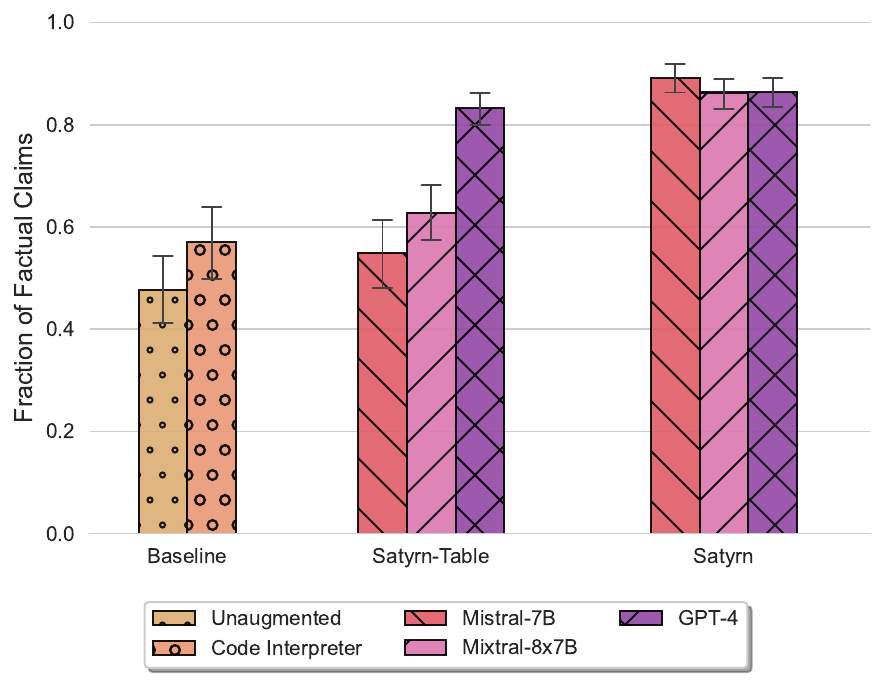}
\caption{The fraction of claims  classified as factual, rather than confabulated or refuted.}
\label{fig:fraction_claims_factual} 
\end{figure}

Figure \ref{fig:fraction_claims_factual} shows the difference in the fraction of claims that were classified as factual for each generation mode. The two baselines, Unaugmented GPT-4 and Code Interpreter, had 48\% and 57\% factual accuracy respectively. The ablated \satyrn{}-Table mode was comparable with both of the baselines, even with the smallest model, Mistral-7B. In this ablation study, we observed that larger models were much better at processing information structured as tables. However, when we structure the information in natural language statements and pass these to the LLMs instead, the differences in accuracy between the small and large models vanish entirely. Mistral-7B benefits the most with an increase of 34 points from its performance in the ablation. \satyrn{} outperforms the closest baseline by 32 points with a much smaller model. 

\subsection{Fluency and Coherence}

We find that reports produced by each generation mode exhibit high degrees of fluency as shown in Table \ref{tab:accuracy_fluency_coherence}. Reports generated by GPT-4 show slightly higher fluency scores than those generated by the smaller models. This is expected behavior since LLMs trained on larger datasets with more parameters are found to exhibit improved generative capabilities \cite{kaplan2020scaling, hoffmann2022empirical}. A similar result is found for the coherence of the reports, with all generation modes demonstrating high levels of coherence as shown in Table \ref{tab:accuracy_fluency_coherence}.

\subsection{Claim Type Breakdown}
\begin{figure}
\centering 
\includegraphics[width=\columnwidth,keepaspectratio]{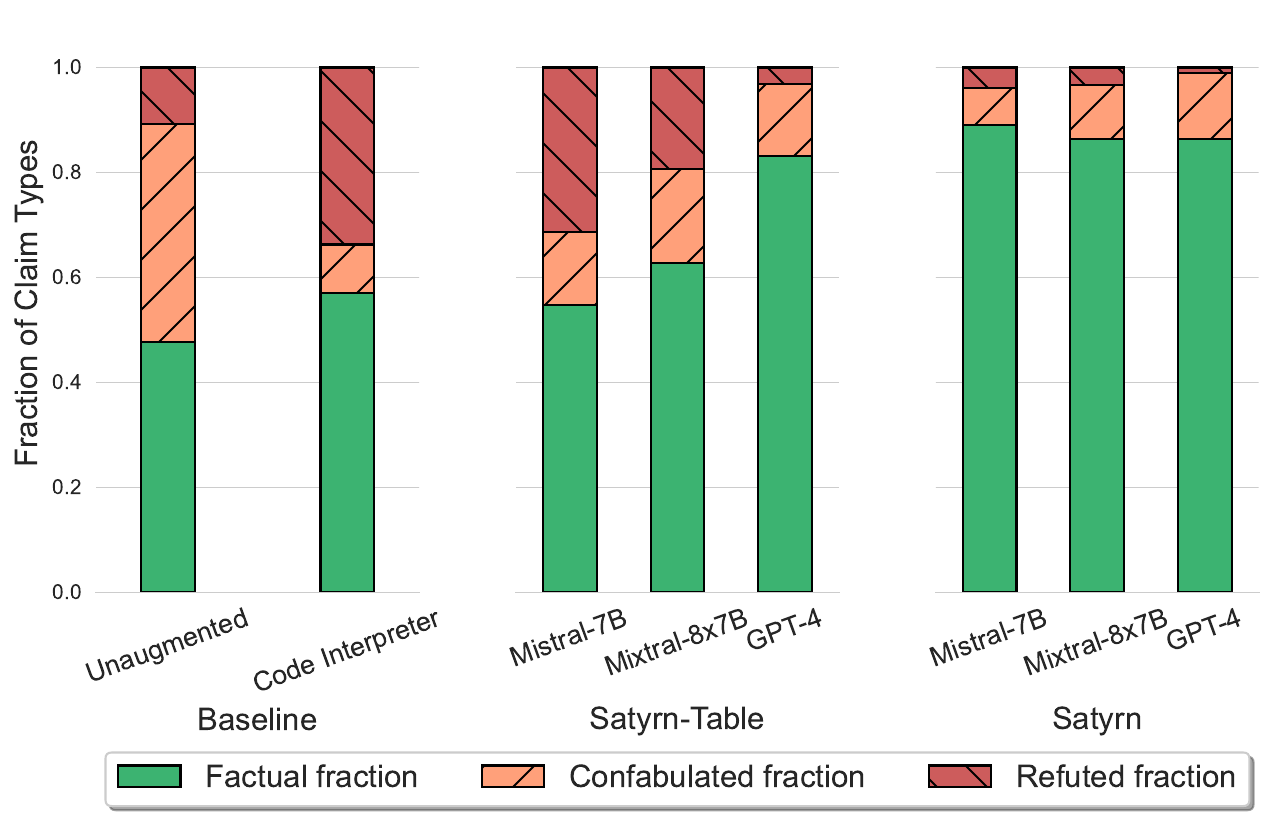}
\caption{The fraction of claims that were classified as factual, confabulated, or refuted.} 
\label{fig:claim_breakdown_small} 
\end{figure}

In Figure \ref{fig:claim_breakdown_small}, we normalize by the total number of claims, and present a breakdown of the fraction of factual, confabulated, and refuted claims. An interesting observation here is the distinction between the types of unsupported claims made by Unaugmented GPT-4 versus Code Interpreter. Unaugmented GPT-4 makes a lot of confabulated claims since it does not have access to any data or knowledge source. It also makes a lot of vague, but true statements (e.g., "This value is higher than the minimum value but not quite at the maximum value."). On the other hand, Code Interpreter is often unable to correctly derive information from the data, resulting in refuted claims.

\begin{figure}
\centering 
\includegraphics[width=\columnwidth,keepaspectratio]{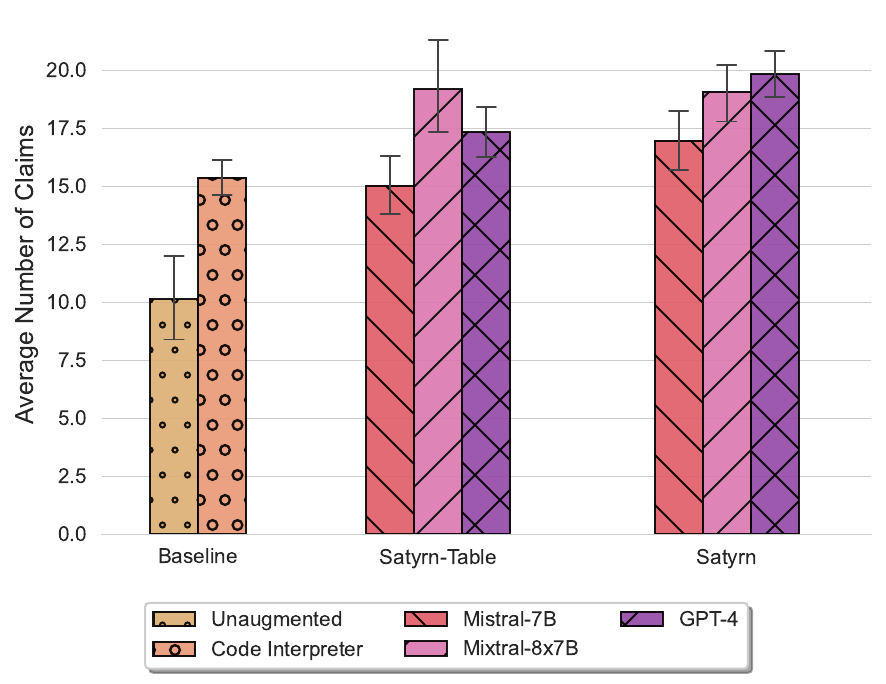}
\caption{The average number of claims made for each configuration across all domains.} 
\label{fig:num_claims} 
\end{figure}

We observe that the Unaugmented GPT-4 baseline has far fewer claims than other modes as seen in Figure \ref{fig:num_claims}. This is due to many of the sentences in its reports containing placeholder values and few concrete claims of fact that can be evaluated (e.g., "In comparison with Y, Bond County ranks Z.").

\section{\satyrn{} Platform}
\label{sec:platform}

\satyrn{} is a platform for enabling users to produce fluent, coherent, and accurate reports that are grounded by their data. \satyrn{}'s capabilities can be expanded with new datasets, new analytics, and new report types.

To utilize a \textbf{new dataset}, a ring can be created. Given a ring, all existing analyses can be applied to the data and all report blueprints can be used to generate reports with them. The only exception is when the data itself does not support a particular kind of report. For example, the Time over Time report used in the experiments requires the data be temporal so that analyses like change over time can be computed. To enable \textbf{new analytics} within Satyrn, additional SQR plan templates can be added to \satyrn{}. To enable a \textbf{new report type} to be generated, a report blueprint can be created. New blueprints can utilize any of the SQR plan templates available within \satyrn{} to deterministically guarantee the desired information is produced when generating a report. This cross-compatibility of all components of the platform ensures that \satyrn{} is scalable to new datasets, new analyses, and new report types.
\section{Related Work}

\textbf{Domain and Analytics Representations} Highly specialized ontologies have been developed for a diverse range of areas such as medicine \cite{salvadores2013bioportal}, law \cite{casellas2011legal}, food \cite{kamel2015towards}, chemical engineering \cite{marquardt2010re}, and biological environments \cite{buttigieg2013environment}. However, the production of ontologies such as these requires extensive expertise in ontology design and substantial amounts of time. \cite{patterson2019teaching} have semantically enriched data science scripts with the goal of successfully modeling computer programs. However, their work focuses more on supporting automated reasoning about data science software rather than encoding core analytic knowledge and processes that can be used when mapping analytics onto data in a domain-agnostic fashion.

\textbf{Knowledge Augmented Generation} One key method to promote factual generation by an LLM is using an external knowledge source to augment the generation \cite{chen-etal-2017-reading, lewis2021retrievalaugmented, shuster2021retrieval, izacard2022atlas, siriwardhana-etal-2023-improving}. Knowledge graphs \cite{min2020knowledge, baek2023knowledge}, textual documents \cite{paranjape2021hindsight, trivedi-etal-2023-interleaving}, pre-processed vectors \cite{verga-etal-2021-adaptable}, search engines \cite{nakano2021webgpt}, and even other LLMs \cite{shwartz-etal-2020-unsupervised} have all been used as external knowledge bases. External symbolic engines have also been used to perform computation or reasoning, the results of which are used to augment an LLM's generation \cite{schick2024toolformer, zhuang2023toolqa, peng2023check}. While our work also utilizes a symbolic engine, we enhance relational databases with lightweight semantics in order to map the data to the operations of this engine and obtain results in natural language.
 
\textbf{Data-to-Text Generation} A related task of data-to-text generation where the goal is to generate descriptions of structured data organized in tables has been studied for a long time \cite{kukich-1983-design, reiter2000nlgbook}. Traditionally, template based algorithms were used to build data-to-text systems \cite{oh-rudnicky-2000-stochastic, stent-etal-2004-trainable, kondadadi-etal-2013-statistical}, while recent approaches have adopted a plan-then-generate procedure \cite{su2021planthengenerate}. Currently, the most popular method is to use end-to-end neural pipelines where the model is fine-tuned to produce text from data \cite{puduppully-etal-2019-data, yang2021table, ghosal-etal-2023-retag, zhao-etal-2023-openrt}. Our approach differs in three key ways: 1) we target large scale databases rather than small tabular data, 2) we use an LLM for generation with no fine-tuning, and 3) our approach enables information not already present in the data to be computed at run-time for use in generation.
\section{Conclusion}

We present \satyrn{}, a system that leverages AAG to produce highly accurate, fluent, and coherent reports that adhere to information derived from data. We find that \satyrn{} generates reports that contain over 86\% accurate claims as compared to GPT-4 Code Interpreter in which just 57\% of claims are accurate. Notably, this is accomplished using a far smaller model, while preserving high levels of fluency and coherence. In the future, we plan to develop automatic validation methods to further improve the accuracy of generated reports. Additionally, we plan to allow language to be used as input rather than structured information requests.
\section*{Limitations}

In this section, we discuss limitations and future directions for this work.

\textbf{Input to \satyrn{} is not language}: The input to \satyrn{} is currently a structured information request containing the type of report to generate, the subject of the report, a metric to use when evaluating the subject, and any filters to apply to the data before analysis. While this structure guarantees \satyrn{} will generate the correct kind of report, it is also more cumbersome to specify than a request in natural language. We plan to address this limitation by utilizing \satyrn{}'s knowledge of the questions it can answer based on the available analytics to build a dataset question and SQR plans pairs in order to train a model for converting questions to SQR plans.

\textbf{Report structure is encoded by model}: One aspect of our report generation method that makes it somewhat limited is our reliance on the LLM to encode the structure of the report type rather than controlling for this ourselves. While we do consider this somewhat desirable, as it prevents us from having to define an explicit structure for each new type of report we wish the system to generate, it also means that we have little control over the document structure and as such, the quality and coherence of the structure depends on how well it is encoded in the LLM.

\textbf{Report validation}: Our approach has another limitation that it shares with RAG approaches: validation. Automatic fact extraction and claim verification is an active area of research \cite{min-etal-2023-factscore, wei2024long, song2024veriscore}. Measuring the factual accuracy of the reports generated in our experiments involved us manually labeling each claim as factual, confabulated or refuted. Further research in the direction will greatly help in evaluating systems like ours, and other natural language generation methods, at a larger scale.
\section*{Ethics Statement}
In this work we present a system that automates the process of generating documents from data. While the aim of \satyrn{} is to ground generated reports in truth, particularly as compared to an unguided LLM, the system does not preclude the generation of reports from factually flawed data. As such, it is contingent on the system's user to verify the accuracy and validity of raw datasets fed to the system.

Additionally, we rely on an LLM to generate the language of a report. Such models have encoded biases that could crop up in the final report and could result in biases or other undesirable language. It is conceivable that bad faith actors could work to manipulate the presentation of information to ultimately misinform people with a highly skewed version of the truth. This remains a challenge for any approach which uses an LLM for generating a document.
\section*{Acknowledgments}

We would like to thank the Center for Advancing the Safety of Machine Intelligence (CASMI) for funding this work. We also thank Mohammed A. Alam, David Demeter, Alexander Einarsson, and Sergio Servantez for providing constructive feedback as well as the ACL ARR reviewers for their valuable comments.

\bibliography{custom}

\appendix

\section{Analysis Engine Implementation}
\label{app:analytics_engine}

\subsection{Plan Parsing and Execution}

Conversion of the graph-structured SQR plan into SQL is done by first breaking the graph into "subplans", where the result of one subplan functions as a data source for subsequent subplans. From each subplan, the necessary information to form an executable query, including entities, their attributes, analytics operations, and filters, is then identified. For example, in the plan in Figure \ref{fig:sqr_example}, the attributes \textit{size} and \textit{year} are retrieved from the \textit{Wildfire} entity while \textit{name} is retrieved from the \textit{State} entity. The \textit{average} operation is applied to \textit{size}, grouped by \textit{name}. A filter is constructed such that each \textit{Wildfire} considered are from the \textit{year} 2000. From this information, the query is constructed using a query abstraction library. For relational databases we use the SQLAlchemy Python package \cite{sqlalchemy}. In this last step, the ring is leveraged to convert the entity and relationship abstractions to the proper tables and joins.

\subsection{SQL Object Relational Mapping}

While \satyrn{}'s analysis engine is designed to be extendable to a variety of data source types, it is currently only configured to execute queries against relational databases. Upon initialization, for each selected ring, a corresponding object relational mapping (ORM) is built using SQLAlchemy. The ORM provides a programmatic interface between the information defined in the ring, and the data stored in a relational database.

The ORM is constructed using configuration mappings defined in the ring, specifically, the tables, columns, and joins between tables. Unlike objects defined in the ring, the ORM objects hold a direct one-to-one correspondence with database objects; ORM entities correspond to tables, attributes correspond to columns, and relationships correspond to groups of joins.

\textbf{Implicit Joins} A major benefit provided by the abstractions of the ring is that, for a given domain, joins between tables need only be defined once -- in the configuration mapping of the ring. No join information of any kind is required in plan definitions. Instead, the system leverages the joins and relationships defined in the ring to determine which SQL joins to use when attributes are selected corresponding to columns of different tables.

All necessary joins between tables within an entity are identified. These intra-entity joins are necessary when multiple attributes are specified as belonging to the same entity, but correspond to columns from different tables. Then, for each pair of entities, relationships are defined by collecting joins along the shortest path connecting the tables of the two entities.

\section{Report Blueprint Implementation}
\label{app:blueprint_implementation}

\satyrn{} provides a platform for defining new types of reports that scope across any domain and dataset. These report types are encoded by a set of report blueprints, and the report generation process uses them to produce the information to be included in the final report. The primary purpose of a report blueprint is to provide the set of core information requirements, defined by a set of SQR plan templates, and rules for retrieving or producing the required \textit{access plans}, filters, and language template fillers that will ultimately be composed with each SQR plan template. This composition process results in an executable SQR plan for each information requirement of the blueprint, and the execution of these plans with the analysis engine produces the raw set of facts used as part of a prompt for generating the final report with a language model.

\subsection{Attribute Access Plans}

When \satyrn{} loads a ring, it generates an \textit{access plan} for each of the attributes: a SQR plan that specifies how to retrieve this attribute from the data. 

\satyrn{} also automatically produces additional attributes by applying analyses to the existing attributes. By combining the set of attributes defined within the ring with the set of available analytics that can be applied to each of them based on their associated attribute types, it is possible to automatically produce an additional set of attributes. For example, if "population" is present as an attribute and has type "Arithmetic", then "average population" would be created since \satyrn{} knows it can apply the average operation to attributes that are "Arithmetic". Each of these derived attributes has an associated access plan as well.

\begin{figure*}[ht]
\centering 
\includegraphics[width=\linewidth,keepaspectratio]{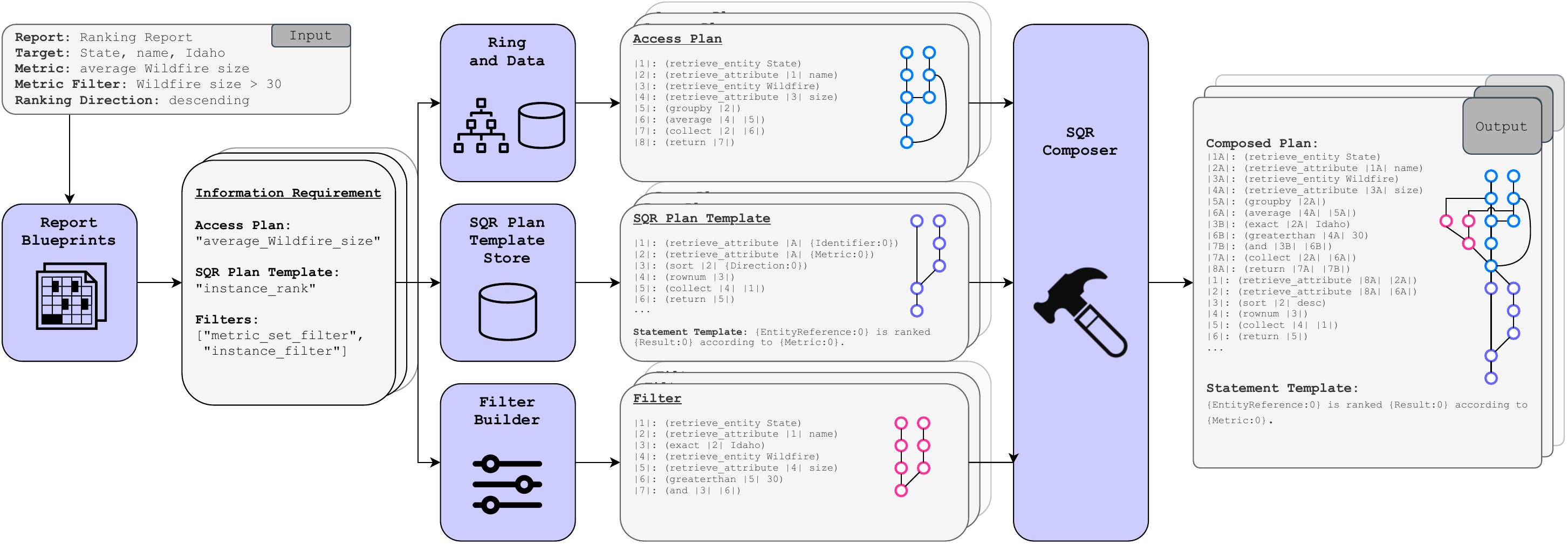}
\caption{This figure shows the process \satyrn{} uses to generate a set of executable SQR plans to derive the information required for a report. First, the specified report type and its requirements are looked up in the set of report blueprints. The SQR plan template associated with each requirement is looked up from the SQR plan template store. Access plans are looked up in the attribute augmented ring. A filter plan is produced and all three components are passed to the SQR composer which assembles them into a single executable plan.} 
\label{fig:sqr_composition_flow} 
\end{figure*}

\subsection{SQR Plan Templates}

The SQR plan templates (previously described in \S \ref{sec:plan_templates}) serve as the base for defining the analytic steps to perform given one or more access plans as input. These input access plans can have filters inserted into them which constrain the retrieval of these data to particular subsets. Each SQR plan template also has a language template associated with it whose slots can be filled via the expression of entities, attributes, and values specified in the SQR plan template.

\subsection{Identifying Plan Composition Inputs}

The inputs to a report blueprint are provided by a user as a JSON object and vary depending on the information requirements of that report. An example of the inputs for a ranking report blueprint can be seen in Figure \ref{fig:sqr_composition_flow}. All currently implemented blueprints require a target entity instance, metric, and metric set filter as part of their input. The user input specifying the target entity instance is used to create an instance filter, which are SQR plan steps used to filter for data relating to this entity instance. The blueprint uses its predefined rules to specify that this filter should be added to particular access plans and designated as input to a SQR plan template. 

This is similarly true for the metric input for the blueprint. If an attribute in the ring has one of its attribute types defined as a Metric, then it is considered a valid input for the blueprint. The metric specified by the user is looked up in the ring, its access plan is retrieved, and then is designated as input to one or more SQR plan templates in the blueprint according to the blueprint rules. The underlying mechanics of how the metric gets mapped to inputs of the SQR plan templates in the blueprint are the same no matter the dataset which is used. There are corresponding rules within the blueprint defining how metric set filters specified by the user are  added to the designated SQR plan templates. Additional report-specific inputs from the user are mapped to the SQR plan templates in a similar fashion. These blueprints ensure \satyrn{}’s production of information from data is non-probabilistic, and a report blueprint is guaranteed to produce the same executable plans for the same user inputs.

\subsection{Executable Plans via SQR Composition}
Ultimately, the rules encoded by the report blueprints, the retrieval of access plans from the ring, the retrieval of SQR plan templates from the SQR plan template store, and the production of filters by the filter builder result in composition specifications. These provide the raw materials for building an executable plan for each of the information requirements. The individual elements comprising the composition specification can be seen as input to the SQR Composer in Figure \ref{fig:sqr_composition_flow}. 

Prior to composition, the SQR Composer ensures that all the references of the access plans, SQR plan template, and filter are unique by appending a unique string to the step references of each SQR plan, preventing conflicts when combining them. Next, the SQR Composer updates the SQR plan template inputs with the correct access plan references (e.g. changing “|A|” to “|8A|” in Figure \ref{fig:sqr_composition_flow}), and finally appends all the steps from the SQR plan template, access plan, and filters together to form an executable plan. \satyrn{} carries out this process for all information requirements of the report blueprint in order to produce a set of executable SQR plans that will be used for deriving the information that should be communicated in the report.

\section{Experimental Details}

\subsection{Report Types}
\label{app:report_types}

We define three report blueprints that focus on common reporting use cases. Their information requirements are shown in Table \ref{tab:blueprint_info_reqs}.

\begin{table*}[ht!]
\small
\centering
\begin{tabular}{p{0.15\linewidth}p{0.75\linewidth}}
\hline
\textbf{Document Type} & \textbf{Information Requirements} \\
\hline
\multirow{7}{*}{Ranking}
& The value of the metric for the target entity \\
\cline{2-2}
& The total number of entities being ranked \\
\cline{2-2}
& The rank of the entity instance according to the metric \\
\cline{2-2}
& The top three entity instances according to the metric \\
\cline{2-2}
& How much lower the target entity is than the highest \\
\cline{2-2}
& The average, minimum, and maximum value of the metric for all entities \\
\cline{2-2}
& Whether or not the target entity is greater than the average value of the metric across all entities \\
\hline
\multirow{5}{*}{\multirow{3}{\linewidth}{Comparative Benchmark}}
& The value of the metric for the target entity \\
\cline{2-2}
& Whether this value is greater than the target benchmark value \\
\cline{2-2}
& What the minimum, maximum, average, and median value of the metric is for all entities \\
\cline{2-2}
& Whether or not the metric value for the target entity is greater than the average value of the metric for all entities \\
\cline{2-2}
& Whether or not the metric value for the target entity is greater than the median value of the metric for all entities  \\
\cline{2-2}
& What the standard deviation of the metric is for all entities \\
\hline
\multirow{5}{*}{Time over Time}
& The values of the metric at the start and end times for the target entity \\
\cline{2-2}
& The percent change in these values \\
\cline{2-2}
& The average, minimum, and maximum value of the metric for all entities at the start and end times \\
\cline{2-2}
& The percent change between of the average between the start and end time for all entities \\
\cline{2-2}
& Whether or not the percent change was greater for the target entity instance or the average for all entities \\
\hline
\end{tabular}
\caption{The information requirements for each of the three report types used in our experiments.}
\label{tab:blueprint_info_reqs}
\end{table*}

\subsection{Hyperparameter Tuning}
When performing the report generation with the language models, no hyperparameter search was performed, and we used default values for all model parameters with the exception of the following. For GPT-4 generations, we used a temperature of 0.0. For Mistral-7B and Mixtral 8x7B, we used a temperature of 0.2 and a top-p of 0.1.

\subsection{Repeated Generation for Code Interpreter}
Code Interpreter would often fail to produce any meaningful outputs or reports when given a prompt to do so. In order to ensure a fair comparison for evaluation, many of the prompts needed to be ran repeatedly in order for a report to be generated.

\subsection{Datasets}
\label{app:datasets}
Table \ref{tab:datasets} lists the 8 datasets with brief descriptions, in the 8 domains used in our experiments.

\begin{table*}[ht!]
\small
\centering
\begin{tabular}{p{0.15\linewidth}p{0.20\linewidth}p{0.55\linewidth}}
\hline
\textbf{Domain} & \textbf{Dataset} & \textbf{Description} \\
\hline
\multirow{3}{\linewidth}{Environmental Sustainability}
& \multirow{3}{*}{Wildfire Occurrence} & The Wildfire Data \cite{USwildfire-2022} provides 2.3 million geo-referenced records on U.S. wildfires from 1992 to 2020, covering 180 million burned acres with key identifiers for data linkage. \\
\hline
\multirow{4}{*}{Healthcare}
& \multirow{4}{*}{MIMIC-IV-ED-Demo} & MIMIC-IV-ED \cite{johnson-2023} is a  deidentified critical care database from BIDMC with 40,000+ patient records, organized modularly for easy access to diverse data sources while complying with HIPAA Safe Harbor. \\
\hline
\multirow{3}{*}{Urban Housing}
& \multirow{3}{\linewidth}{Zillow Observed Rent Index} & Zillow Observed Rent Index \cite{zillow-group-inc-2023} is a representative dollar-denominated rental index, calculated from listed rents in the 40th to 60th percentile for all housing types in various regions. \\
\hline
\multirow{4}{*}{Criminal Justice}
& \multirow{4}{\linewidth}{School Shooting Incidents} & The dataset \cite{center-for-homeland-defense-and-security-2023} covers publicly available data on shooting incidents from 1970 to June 2022, including any instance of gun brandishing, firing, or bullets hitting school property, regardless of outcomes or timing. \\
\hline
\multirow{3}{*}{Education}
& \multirow{3}{*}{Illinois Report Card} & The Illinois Report Card \cite{jmlarkin-2020}, issued by the Illinois State Board of Education, provides annual educational progress data for the state, schools, and districts. \\
\hline
\multirow{4}{*}{Legal and Judicial}
& \multirow{4}{*}{SCALES} & The SCALES dataset \cite{paley2021data} combines data from PACER, including ten years of docket reports (2007-2016) from Northern Illinois district courts and 2016 district court reports, with the Federal Judicial Center's judge metadata. \\
\hline
\multirow{5}{*}{Socioeconomic}
& \multirow{5}{*}{Income Disparity} & The U.S. Bureau of Economic Analysis' report \cite{us-bureau-of-economic-analysis-personal-income-2022,us-bureau-of-labor-statistics-unemployed-2023, us-census-bureau-age-17-2022, us-census-bureau-median-household-2022, us-census-bureau-all-age-2022} captures personal income data for various regions, showing income received by residents in those areas based on their place of residence. \\
\hline
\multirow{3}{*}{Business}
& \multirow{3}{*}{Yelp Open Dataset} & The Yelp Open Dataset \cite{yelp-no-date} comprises 5.9 million reviews, 188,593 businesses, and 280,992 pictures, provided by Yelp for personal, educational, and academic purposes. \\
\hline

\end{tabular}
\caption{The eight datasets used in our experiments.}
\label{tab:datasets}
\end{table*}

\section{List of SQR Operations}
\label{app:sqr_operations}

A listing of analytic operations that make up the SQR primitives can be seen in Table \ref{tab:analytic_operations}. This set of operations is easily extensible to new analytics.

\begin{table*}[ht!]
\small
\centering
\begin{tabular}{p{0.15\linewidth}p{0.12\linewidth}p{0.05\linewidth}p{0.25\linewidth}p{0.05\linewidth}p{0.25\linewidth}}
\hline
\textbf{Operation} & \textbf{Operation Type}& \textbf{Input Arity} & \textbf{Input Attribute Types}& \textbf{Output Arity} &\textbf{Output Attribute Types} \\
\hline

\multirow{2}{*}{Average}
& \multirow{2}{*}{Aggregation} & $1$ & [Arithmetic, Metric] & $1$ & [Arithmetic, Metric]  \\
&  & $\leq1$ & [Grouping] &  \\
\hline

\multirow{2}{*}{Correlation}
& \multirow{2}{*}{Aggregation} & $2$ & [Arithmetic, Metric, Datetime] & $1$ & [Arithmetic, Metric, Datetime]  \\
&  & $\leq1$ & [Grouping] &  \\
\hline

\multirow{2}{*}{Count}
& \multirow{2}{*}{Aggregation} & $1$ & [Arithmetic, Metric] & $1$ & [Arithmetic, Metric]  \\
&  & $\leq1$ & [Grouping] &  \\
\hline

\multirow{2}{*}{Count Unique}
& \multirow{2}{*}{Aggregation} & $1$ & [Arithmetic, Metric] & $1$ & [Arithmetic, Metric]  \\
&  & $\leq1$ & [Grouping] &  \\
\hline

\multirow{2}{*}{Get One}
& \multirow{2}{*}{Aggregation} & $1$ & [Arithmetic, Metric, Datetime] & $1$ & [Arithmetic, Metric, Datetime]  \\
&  & $\leq1$ & [Grouping] &  \\
\hline

\multirow{2}{*}{Max}
& \multirow{2}{*}{Aggregation} & $1$ & [Arithmetic, Metric, Datetime] & $1$ & [Arithmetic, Metric, Datetime]  \\
&  & $\leq1$ & [Grouping] &  \\
\hline

\multirow{2}{*}{Median}
& \multirow{2}{*}{Aggregation} & $1$ & [Arithmetic, Metric, Datetime] & $1$ & [Arithmetic, Metric, Datetime]  \\
&  & $\leq1$ & [Grouping] &  \\
\hline

\multirow{2}{*}{Min}
& \multirow{2}{*}{Aggregation} & $1$ & [Arithmetic, Metric, Datetime] & $1$ & [Arithmetic, Metric, Datetime]  \\
&  & $\leq1$ & [Grouping] &  \\
\hline

\multirow{2}{*}{Standard Deviation}
& \multirow{2}{*}{Aggregation} & $1$ & [Arithmetic, Metric] & $1$ & [Arithmetic, Metric]  \\
&  & $\leq1$ & [Grouping] &  \\
\hline

\multirow{2}{*}{String Aggregation}
& \multirow{2}{*}{Aggregation} & $1$ & [Arithmetic, Metric, Datetime] & $1$ & [Arithmetic, Metric, Datetime]  \\
&  & $\leq1$ & [Grouping] &  \\
\hline

\multirow{2}{*}{Sum}
& \multirow{2}{*}{Aggregation} & $1$ & [Arithmetic] & $1$ & [Arithmetic, Metric]  \\
&  & $\leq1$ & [Grouping] &  \\
\hline

And & Boolean & $\geq1$ & [Filter] & $1$ & [Filter] \\
\hline

\multirow{2}{*}{Contains}
& \multirow{2}{*}{Boolean} & $1$ & [Attribute] & $1$ & [Filter] \\
&  & $1$ & [Metric] &  \\
\hline

Exact & Boolean & $2$ & [Arithmetic, Metric, Categorical, String, Datetime, Identifier] & $1$ & [Filter] \\
\hline

Greater Than & Boolean & $2$ & [Arithmetic, Metric, Datetime] & $1$ & [Filter] \\
\hline

Grtr. Than Equal & Boolean & $2$ & [Arithmetic, Metric, Datetime] & $1$ & [Filter] \\
\hline

Less Than & Boolean & $2$ & [Arithmetic, Metric, Datetime] & $1$ & [Filter] \\
\hline

Less Than Equal & Boolean & $2$ & [Arithmetic, Metric, Datetime] & $1$ & [Filter] \\
\hline

Not & Boolean & $1$ & [Filter] & $1$ & [Filter] \\
\hline

Or & Boolean & $\geq1$ & [Filter] & $1$ & [Filter] \\
\hline

Absolute Value & Arithmetic & $1$ & [Arithmetic, Metric] & $1$ & [Arithmetic, Metric] \\
\hline

Add & Arithmetic & $\geq2$ & [Arithmetic, Metric] & $1$ & [Arithmetic, Metric, Datetime] \\
\hline

Divide & Arithmetic & $\geq2$ & [Arithmetic, Metric] & $1$ & [Arithmetic, Metric, Datetime] \\
\hline

\multirow{2}{*}{Duration}
& \multirow{2}{*}{Arithmetic} & $1$ & [Datetime] & $1$ & [Arithmetic, Metric] \\
&  & $1$ & [Datetime] &  \\
\hline

Multiply & Arithmetic & $\geq2$ & [Arithmetic, Metric] & $1$ & [Arithmetic, Metric, Datetime] \\
\hline

Percent Change & Arithmetic & $2$ & [Arithmetic, Metric] & $1$ & [Arithmetic, Metric] \\
\hline

Square Root & Arithmetic & $1$ & [Arithmetic, Metric, Datetime] & $1$ & [Arithmetic, Metric, Datetime] \\
\hline

Subtract & Arithmetic & $\geq2$ & [Arithmetic, Metric, Datetime] & $1$ & [Arithmetic, Metric, Datetime] \\
\hline

Collect & Data Operation & $\geq1$ & [Attribute] & $1$ & [AttributeCollection] \\
\hline

Groupby & Data Operation & $\geq1$ & [Categorical, Datetime] & $1$ & [Grouping] \\
\hline

Limit & Data Operation & $1$ & [Attribute] & $1$ & [Limit] \\
\hline

\multirow{4}{*}{Return}
& \multirow{4}{*}{Data Operation} & $1$ & [AttributeCollection] & $1$ & [Entity]  \\
&  & $\leq1$ & [Filter] &  \\
&  & $\leq1$ & [Sort] &  \\
&  & $\leq1$ & [Limit] &  \\
\hline

Row Number & Data Operation & $1$ & [Sort] & $1$ & [RowNum] \\
\hline

\multirow{2}{*}{Sort}
& \multirow{2}{*}{Data Operation} & $\geq1$ & [Attribute] & $1$ & [Sort] \\
&  & $1$ & [String] &  \\
\hline

\multirow{2}{*}{Retrieve Attribute}
& \multirow{2}{*}{Retrieval} & $1$ & [Entity] & $1$ & [Attribute] \\
&  & $1$ & [String] &  \\
\hline

Retrieve Entity & Retrieval & $1$ & [String] & $1$ & [Entity] \\
\hline

\end{tabular}
\caption{The set of operations currently implemented and used within SQR plans. This includes operations used for retrieval, analysis, filtering, and data transformations.}
\label{tab:analytic_operations}
\end{table*}

\section{Accuracy Evaluation Details}
\label{app:accuracy_evaluation}

In this section we present the definition of a claim (\S \ref{app:claim_identification}), the types these claims can be classified as (\S \ref{app:claim_classification}), and the full rubric used during the evaluation of each report's factual accuracy (\S \ref{app:evaluation_rubric}).

\subsection{Claim Identification}
\label{app:claim_identification}

We define a claim as any assertion of truth involving some retrieval or analytic processing of information from the data. Interstitial writing that provides a transition between content, titles, and broad introductions to the report are not considered claims. We manually examine each generated report and identify the claims being made.

\subsection{Claim Classification}
\label{app:claim_classification}

Once a claim has been identified, we determine which of the following mutually exclusive categories this claim belongs to based on the factual statements generated for that kind of report. 

\textbf{Factual}: The claim is directly stated by a fact in the context, or the claim can be directly inferred from the facts in the context via an unambiguous analytic process.

\textbf{Refuted}: The claim is directly refuted by a statement in the context, or the claim could be directly inferred from the facts in the context via an unambiguous analytic process, but is incorrect.

\textbf{Confabulation}: The claim is not directly pulled from the facts in the prompt, or is not inferred from the facts in the prompt.

\subsection{Claim Evaluation Rubric}
\label{app:evaluation_rubric}

\begin{figure*}[ht]
\centering 
\fbox{\includegraphics[width=\linewidth,keepaspectratio]{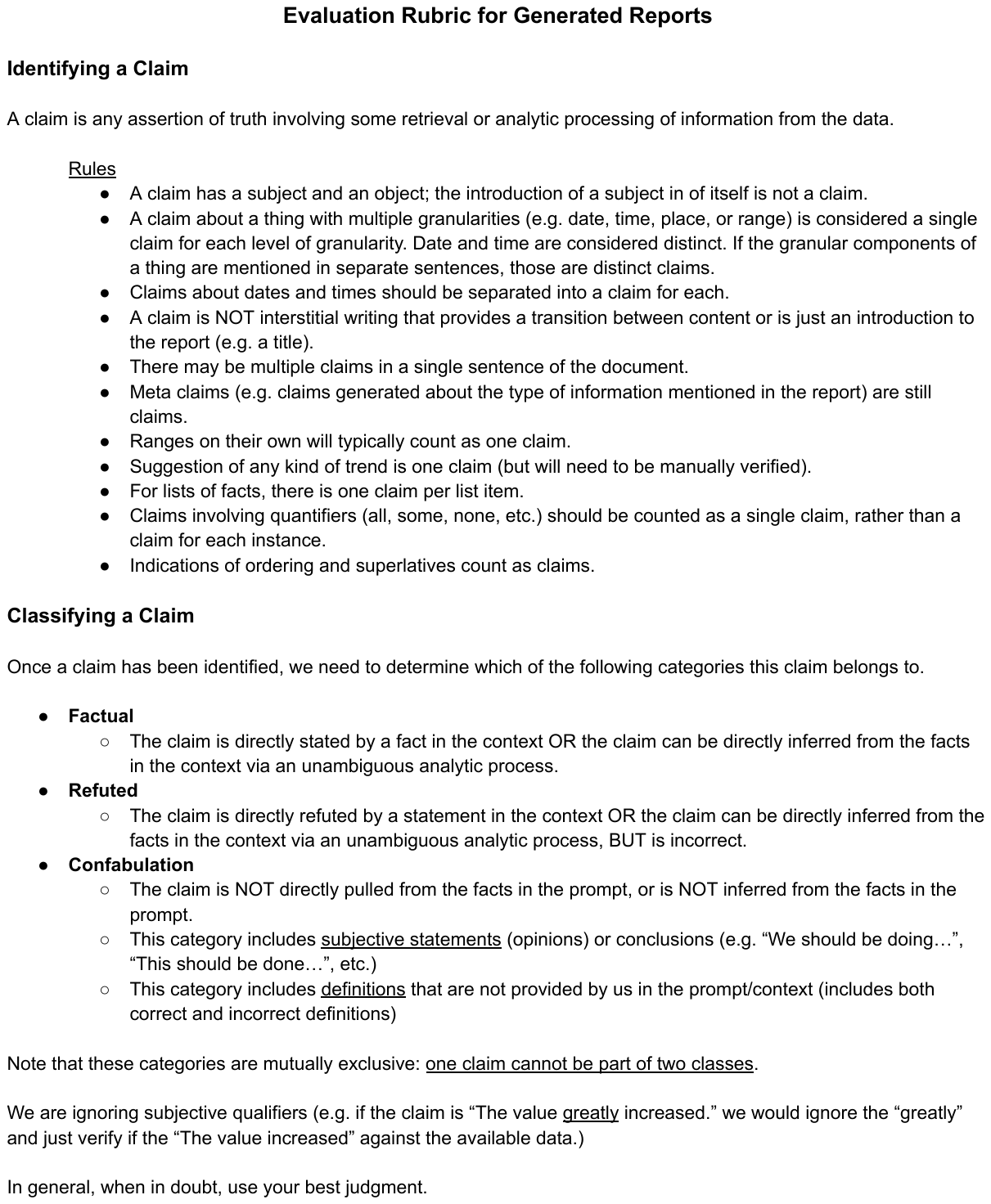}}
\caption{A copy of the rubric that was used for evaluating the reports. This includes both the identification of claims as well as their classification.} 
\label{fig:evaluation_rubric} 
\end{figure*}

A copy of the rubric that was used for evaluation can be seen in Figure \ref{fig:evaluation_rubric}. It includes the instructions used for identifying claims made in a report as well as classifying those claims as factual, refuted, or confabulation.

Note that the confabulation category includes subjective statements or conclusion (e.g. "We should be doing...", "This should be done...", etc.) and definitional statements that are not provided to the language model in the context. It is important to note that not all confabulations are bad. If the model were to define standard deviation in the document, this can be a useful piece of information. However, we do not want to rely on the model to present this, especially if these definitions are inadequate or imprecise. Rather, this is an indication that we should seek to present this definition directly to the model. 

\section{Examples of Generated Reports}
\label{app:example_docs}

\subsection{Ranking Report across Multiple Domains}

\begin{figure*}[ht]
\centering 
\fbox{\includegraphics[width=\linewidth,keepaspectratio]{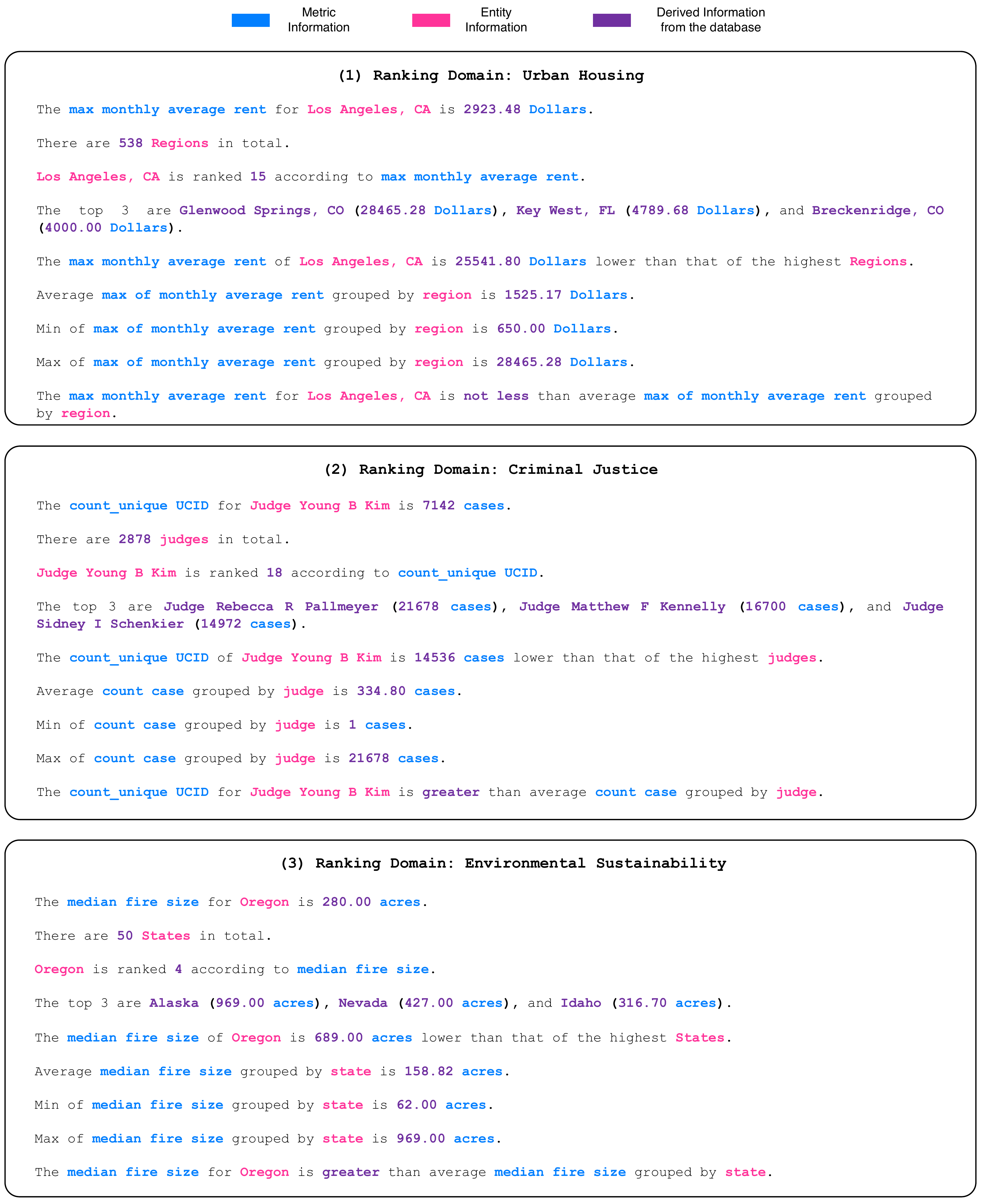}}
\caption{Factual statements generated across three domains for the same ranking blueprint.} 
\label{fig:cross_domain_statements} 
\end{figure*}

We present an example of the domain-agnostic usage of report blueprints. Figure \ref{fig:cross_domain_statements} showcases the factual statements generated by the same \textbf{Ranking} report blueprint, on three different domains: urban housing, criminal justice, and environmental sustainability. The information requirements of the Ranking report blueprint remain unchanged across domains. The target entity information and the metric are specific to the domain and are provided as inputs to the blueprint. Figure \ref{fig:cross_domain_reports} presents the corresponding reports generated by the language model when prompted with the corresponding instruction and factual statements from Figure \ref{fig:cross_domain_statements}.

\begin{figure*}[ht]
\centering 
\fbox{\includegraphics[width=\linewidth,keepaspectratio]{figures/cross_domain_reports.pdf}}
\caption{Ranking Reports generated across three domains using generalized blueprint.} 
\label{fig:cross_domain_reports} 
\end{figure*}

\subsection{Report from All Generation Modes}

We present examples of the same Time over Time report in the Socioeconomics domain generated with various combinations of prompt types and language models. All examples shown in this section are generated for the same report. The target entity is Lake County, IL, the metric is average percent of people in poverty, and the dataset is filtered to only include counties with resident populations greater than 100,000. This is a Time over Time report for the time period ranging from 2010 to 2020.

For each example, we provide the report and the prompt that was used to generate it. For clarity, we have color coordinated the facts provided in the prompt with the corresponding text that was generated as part of the reports to enable quick and direct comparisons between the facts provided to the model and its outputs.

\begin{figure*}[ht]
\centering 
\fbox{\includegraphics[width=\linewidth,keepaspectratio]{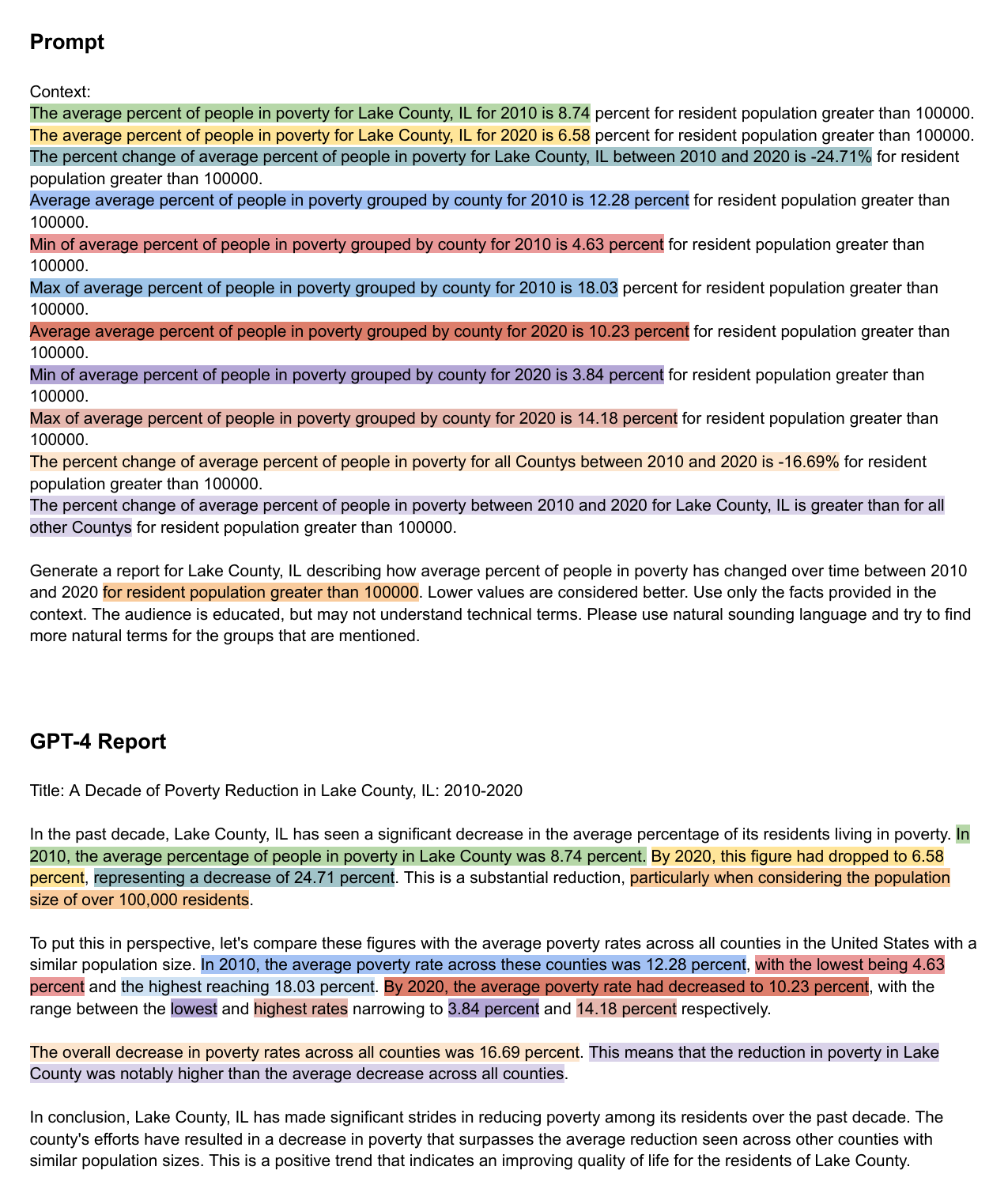}}
\caption{A Time over Time report generated with \satyrn{} using GPT-4 and facts provided as natural language statements.} 
\label{fig:satyrn_gpt4_template} 
\end{figure*}

The first report, shown in Figure \ref{fig:satyrn_gpt4_template}, was generated by \satyrn{} using GPT-4 and facts provided as natural language statements.

\begin{figure*}[ht]
\centering 
\fbox{\includegraphics[width=\linewidth,keepaspectratio]{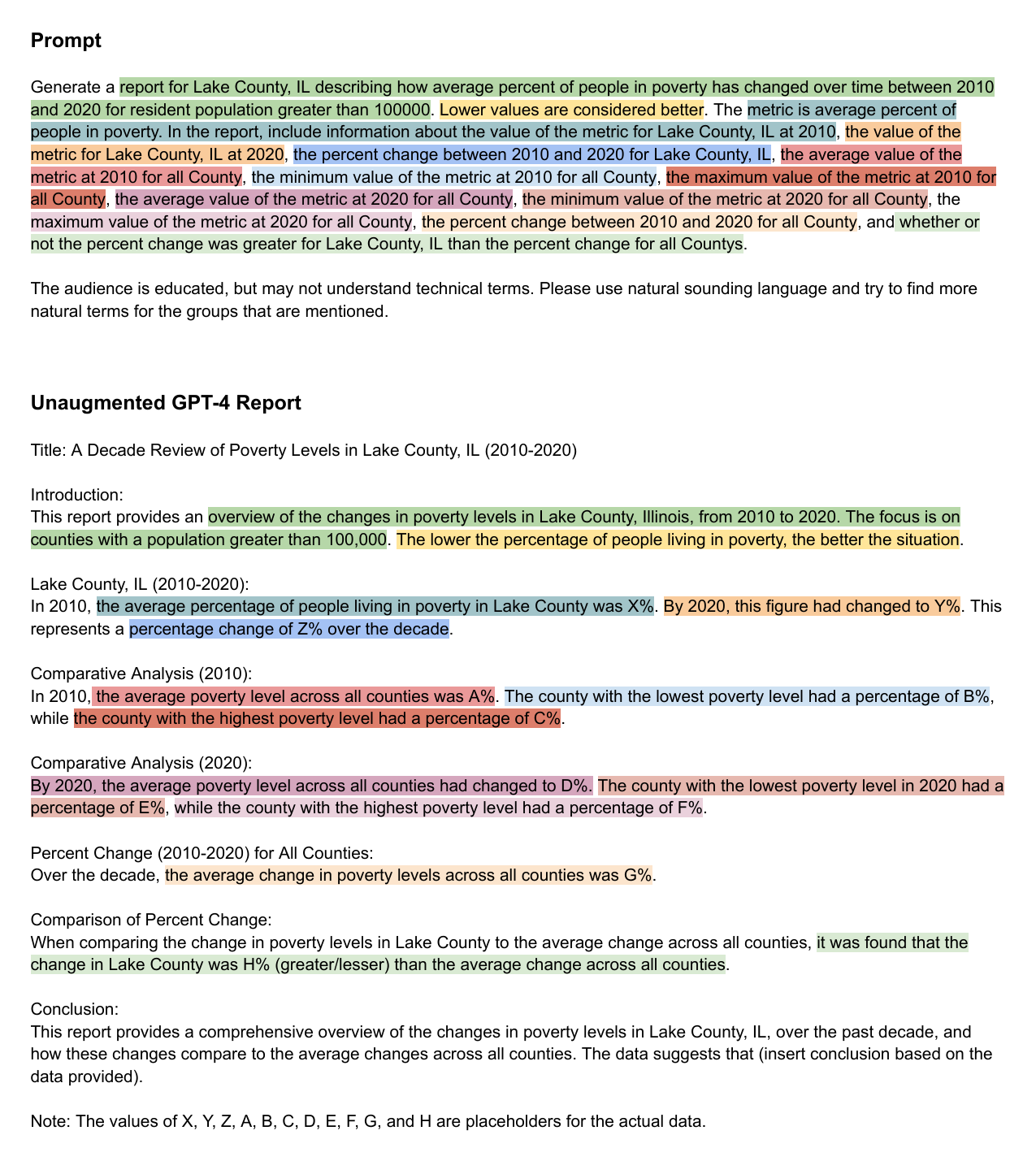}}
\caption{A Time over Time report generated with the Unaugmented GPT-4 baseline. Note that the color coding here is used to highlight where the information should have been if no placeholder values were inserted by the model. This contrasts with the other example outputs which have actual claims with values.} 
\label{fig:no_augmentation_gpt4} 
\end{figure*}

The second report, shown in Figure \ref{fig:no_augmentation_gpt4}, was generated with the Unaugmented GPT-4 baseline. It was observed that Unaugmented baseline reports had a placeholder for the values which can be used if correct data is provided.

\begin{figure*}[ht]
\centering 
\fbox{\includegraphics[width=\linewidth,keepaspectratio]{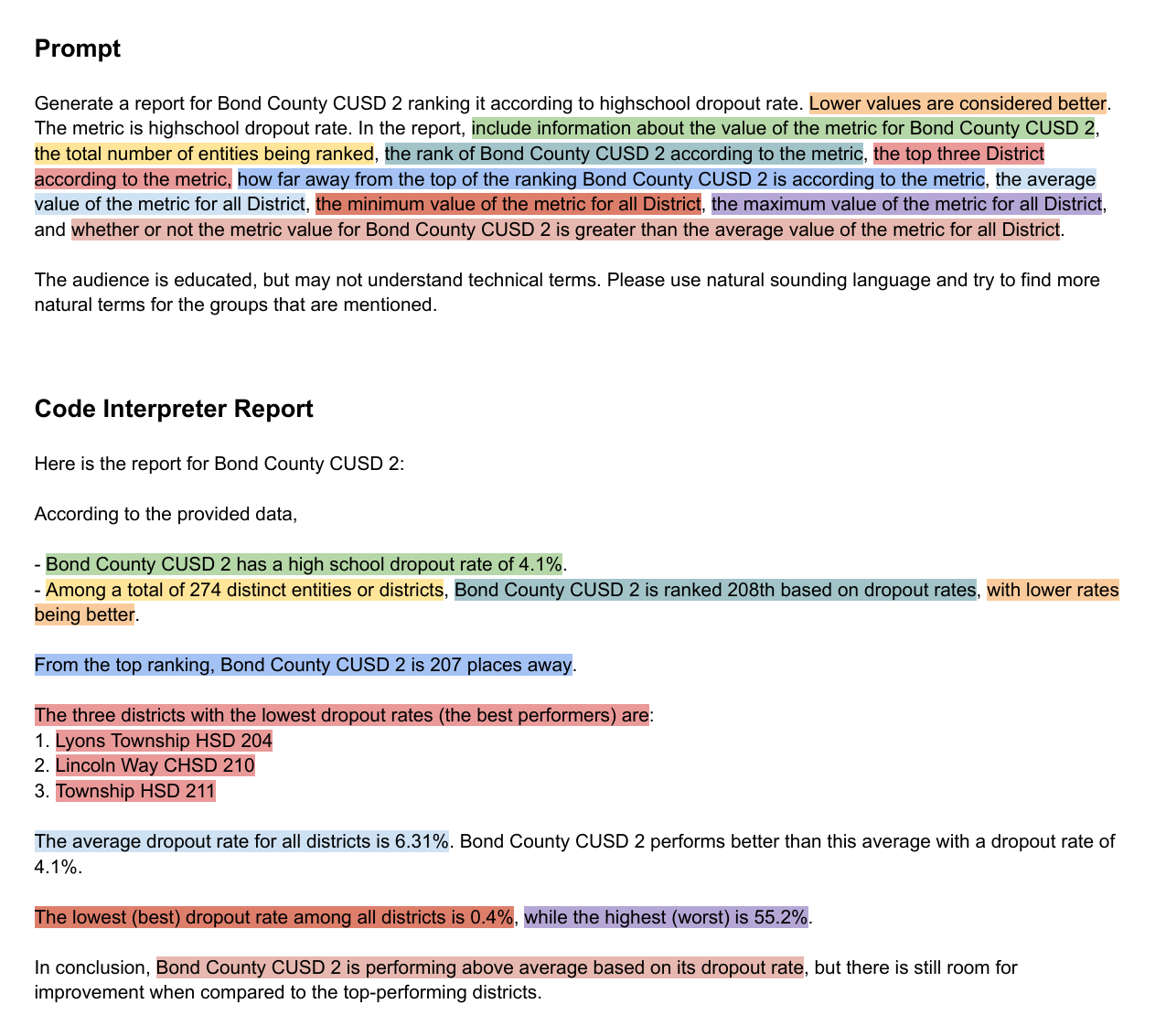}}
\caption{A Time over Time report generated with GPT-4 Code Interpreter with access to the data} 
\label{fig:code_interpreter_gpt4} 
\end{figure*}

The third report, shown in Figure \ref{fig:code_interpreter_gpt4}, was generated by Code Interpreter using GPT-4 with access to the data. Notice that in the report, Code Interpreter is able to come up with values for the desired analysis, but is not correct all the time.

\begin{figure*}[ht]
\centering 
\fbox{\includegraphics[width=\linewidth,keepaspectratio]{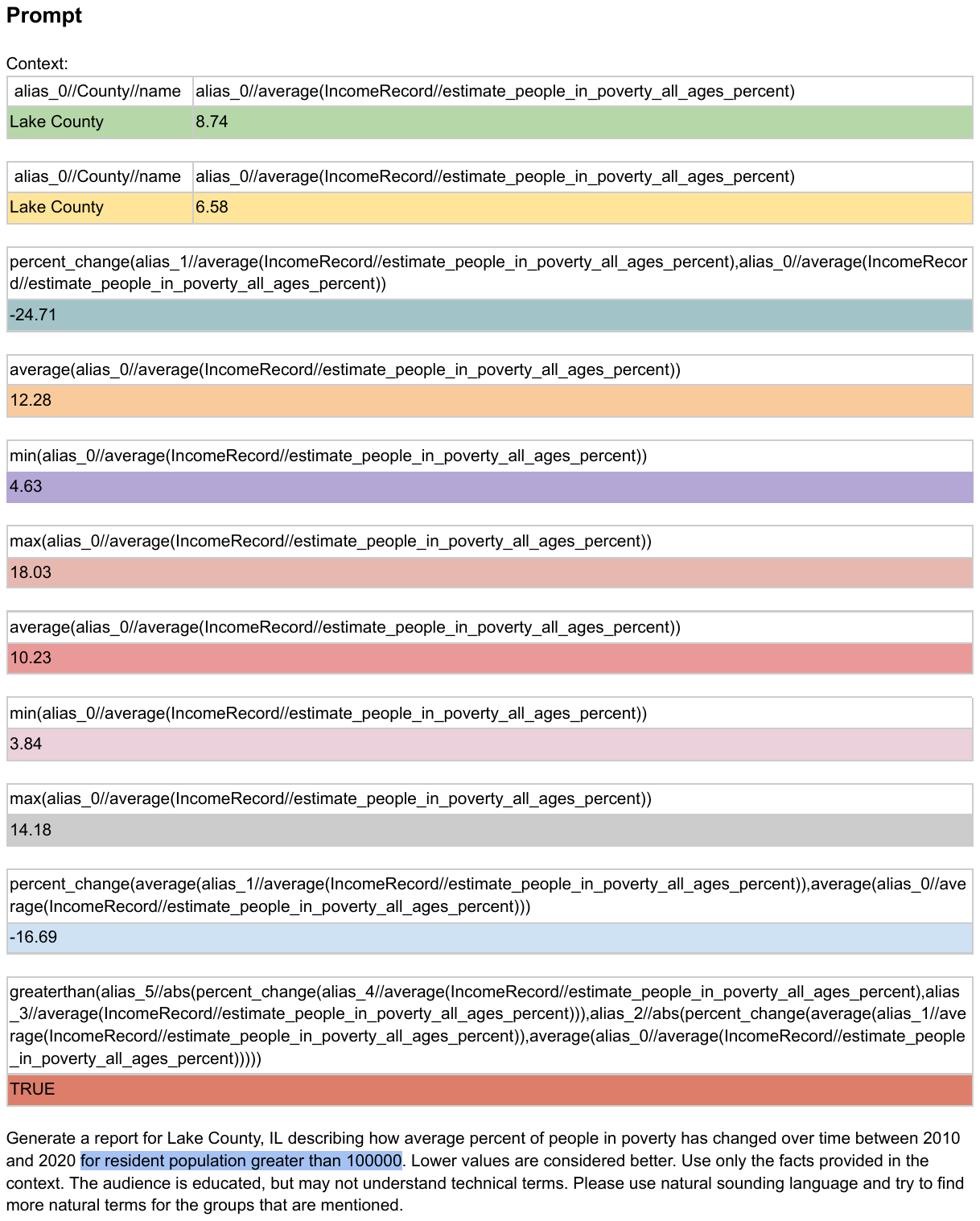}}
\caption{A prompt used for generating a Time over Time report with \satyrn{} where the facts are provided as tables rather than natural language statements.} 
\label{fig:satyrn_table_prompt} 
\end{figure*}

\begin{figure*}[ht]
\centering 
\fbox{\includegraphics[width=\linewidth,keepaspectratio]{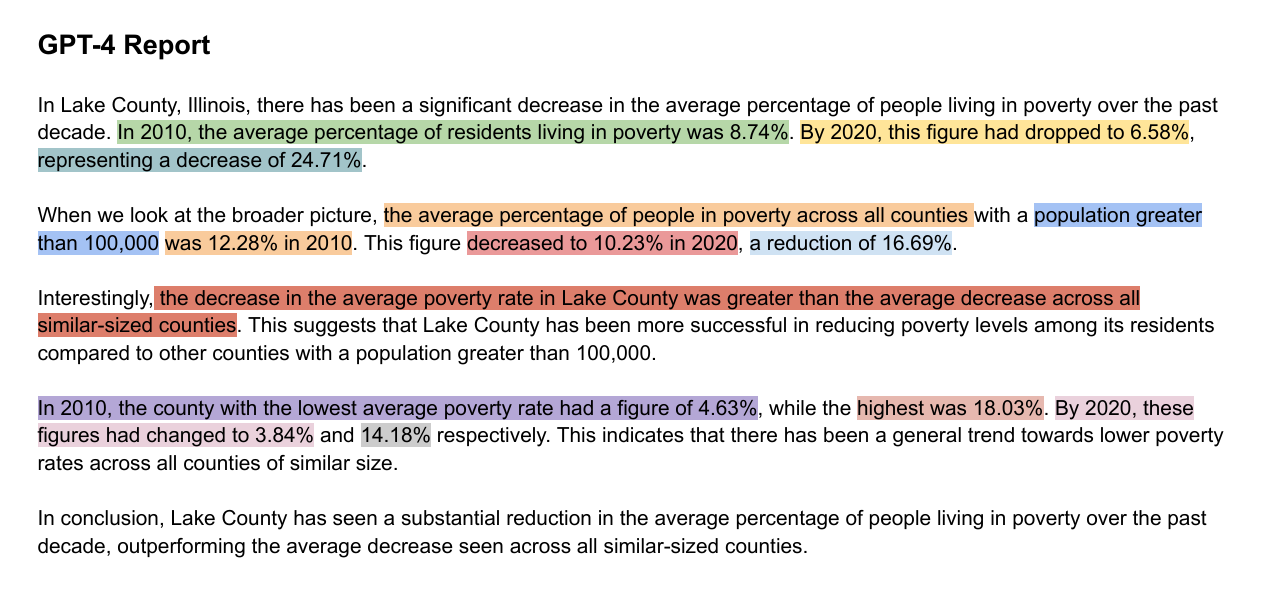}}
\caption{A Time over Time report generated with \satyrn{} using GPT-4 and facts provided as tables rather than natural language statements.} 
\label{fig:satyrn_gpt4_table} 
\end{figure*}

The fourth report, shown in Figure \ref{fig:satyrn_gpt4_table}, was generated by \satyrn{} using GPT-4 with facts provided as tables rather than natural language statements. The prompt used to generate it is shown in Figure \ref{fig:satyrn_table_prompt}.

\begin{figure*}[ht]
\centering 
\fbox{\includegraphics[width=\linewidth,keepaspectratio]{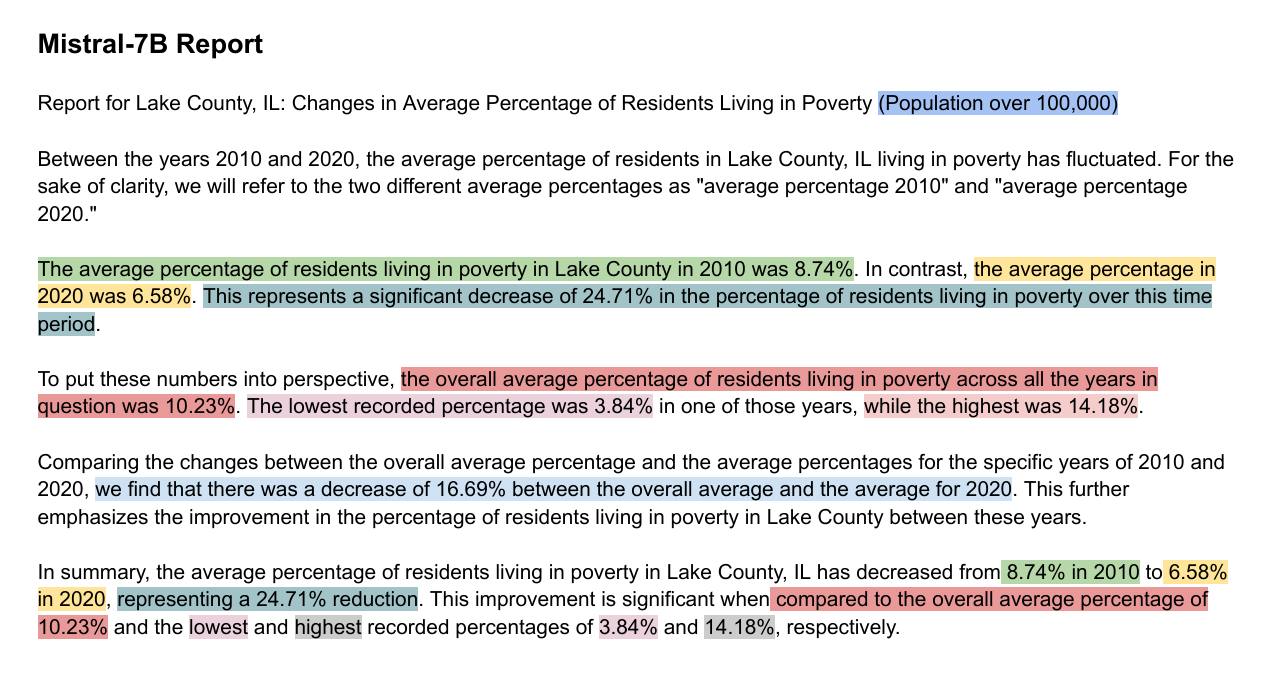}}
\caption{A Time over Time report generated with \satyrn{} using Mistral-7B and facts provided as tables rather than natural language statements.} 
\label{fig:satyrn_mistral_table} 
\end{figure*}

\begin{figure*}[ht]
\centering 
\fbox{\includegraphics[width=\linewidth,keepaspectratio]{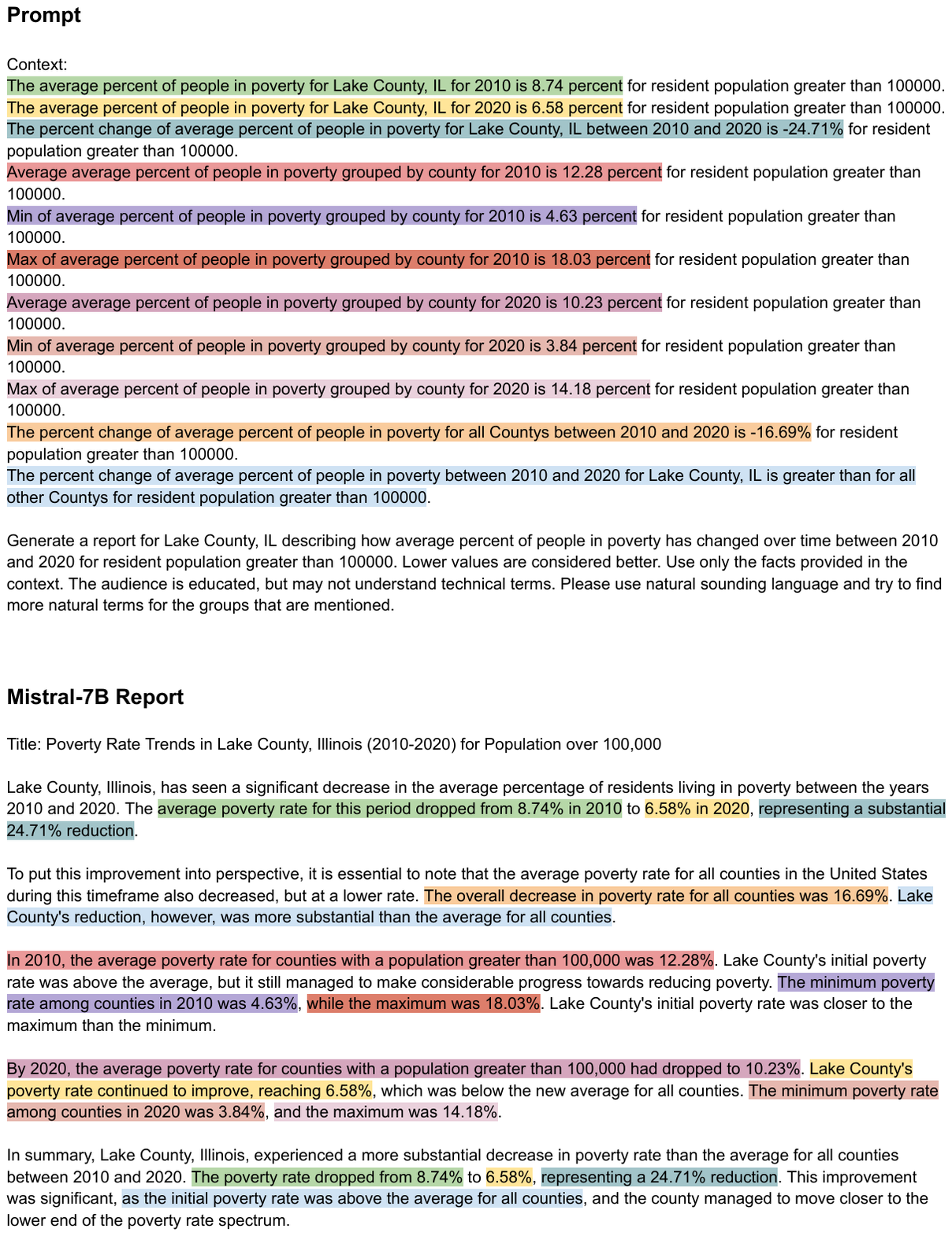}}
\caption{A Time over Time report generated with \satyrn{} using Mistral-7B and facts provided as natural language statements.} 
\label{fig:satyrn_mistral_template} 
\end{figure*}

The fifth report, shown in Figure \ref{fig:satyrn_mistral_table}, was generated by \satyrn{} using Mistral-7B and facts provided as tables rather than natural language statements. The prompt used to generate it is shown in Figure \ref{fig:satyrn_table_prompt}. With Mistral-7B, we observed that the table format was not always understood by the model and the language was sometimes confusing. Whereas Mistral-7B performed well in generating reports when prompts included facts provided as natural language statements (Figure \ref{fig:satyrn_mistral_template}.

It is worth noting that, in general, the models' generations presented the facts in the same order in which they were provided within the prompt. This property could likely be exploited to improve the structure of the reports by having a planning module determine what information is best seen first. We leave this for future work.

\end{document}